\documentclass[10pt,twocolumn,letterpaper]{article}

\usepackage{iccv}
\usepackage{times}
\usepackage{epsfig}
\usepackage{graphicx}
\usepackage{amsmath}
\usepackage{amssymb}

\usepackage{booktabs}
\usepackage{lipsum}
\usepackage{subcaption}
\usepackage{breqn}
\usepackage{times}
\usepackage{textcomp}
\usepackage{bm}
\usepackage{mathtools}

\DeclarePairedDelimiter\floor{\lfloor}{\rfloor}
\usepackage{dblfloatfix}
\usepackage[font=small]{caption} 
\usepackage[export]{adjustbox}

\edef\restoreparindent{\parindent=\the\parindent\relax}
\usepackage{parskip}
\restoreparindent
\usepackage[pagebackref=true,breaklinks=true,letterpaper=true,colorlinks,bookmarks=false]{hyperref}

\iccvfinalcopy 

\def\iccvPaperID{7844} 

\begin{document}


\title{Dissecting Image Crops}

\author{Basile Van Hoorick\\
Columbia University\\
New York, NY, USA\\
{\tt\small basile@cs.columbia.edu}
\and
Carl Vondrick\\
Columbia University\\
New York, NY, USA\\
{\tt\small vondrick@cs.columbia.edu}
}

\maketitle
\ificcvfinal\thispagestyle{empty}\fi


\begin{abstract}
    The elementary operation of cropping underpins nearly every computer vision system, ranging from data augmentation and translation invariance to computational photography and representation learning. This paper investigates the subtle traces introduced by this operation. For example, despite refinements to camera optics, lenses will leave behind certain clues, notably chromatic aberration and vignetting. Photographers also leave behind other clues relating to image aesthetics and scene composition. We study how to detect these traces, and investigate the impact that cropping has on the image distribution.
    While our aim is to dissect the fundamental impact of spatial crops, there are also a number of practical implications to our work,
    such as revealing faulty photojournalism and equipping neural network researchers with a better understanding of shortcut learning.
    Code is available at {\footnotesize\url{https://github.com/basilevh/dissecting-image-crops}}.
    \end{abstract}


\vspace{-1em}
\section{Introduction}

The basic operation of cropping an image underpins nearly every computer vision paper that you will be reading this week.
Within the first few lectures of most introductory computer vision courses, convolutions are motivated as enabling feature invariance to spatial shifts and cropping \cite{brown_course_cv, stanford_course_cv, mit_course_cv}. Neural networks rely on image crops as a form of data augmentation \cite{krizhevsky2017imagenet, szegedy2015going, he2016deep}. Computational photography applications will automatically crop photos in order to improve their aesthetics \cite{samii2015data, chen2016automatic, zeng2019reliable}.
Predictive models extrapolate pixels out from crops \cite{teterwak2019boundless, wang2019wide, van2019image}.
Even the latest self-supervised efforts depend on crops for contrastive learning to induce rich visual representations \cite{chen2020simple,he2020momentum,pathak2016context,selvaraju2020casting}.

\begin{figure}
    \centering
    \includegraphics[width=\linewidth]{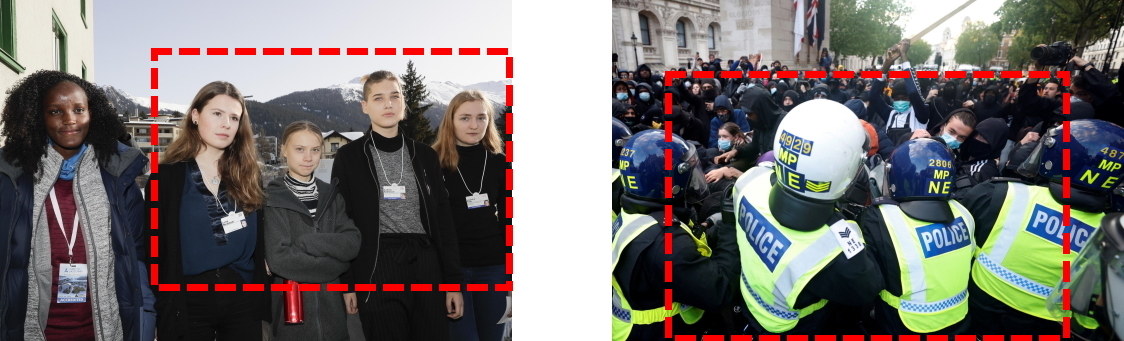}
    \caption{
    We show two infamous image crops, visualized by the red box.
    (\textbf{left}) An Ugandan climate activist had been cropped out of the photo before it was posted in an online news article, the discovery of which sparked controversy \cite{esfandiari_martin_2020}. (\textbf{right}) A news network had cropped out a large stick being held by a demonstrator during a protest \cite{sherling_2020}.  Cropping dramatically alters the message of the photographs.
    }
    \label{fig:teaser_answer}
    \vspace{-1em}
\end{figure}

This core visual operation can have a significant impact on photographs. As Oliva and Torralba told us twenty years ago, scene context drives perception \cite{oliva2001modeling}. Recently, image cropping has been at the heart of media disinformation.
Figure \ref{fig:teaser_answer} shows two popular photographs where the photographer or media organization spatially cropped out part of the context, altering the message of the image.
Twitter's auto-crop feature relied on a saliency prediction network that was racially biased \cite{canales_2020}.

The guiding question of this paper is to understand the traces left behind from this fundamental operation. What impact does image cropping have on the visual distribution? Can we determine when and how a photo has been cropped?

Despite extensive refinements to the manufacturing process of camera optics and sensors, nearly every modern camera pipeline will leave behind subtle lens artefacts onto the photos that it captures. For example, vignetting is caused by a lens focusing more light at the center of the sensor, creating images that are slightly brighter in the middle than near its borders \cite{lopez2015revisiting}. Chromatic aberration, also known as purple fringing, is caused by the lens focusing each wave length differently \cite{beeson2007patterns}. Since these artefacts are correlated with their spatial position in the image plane, they cause image crops to have trace signatures. 

Physical aberrations are not the only traces left behind during the operation. Photographers will prefer to take photos of interesting objects and in canonical poses \cite{torralba2011unbiased, barbu2019objectnet, hendrycks2019natural}.
Aesthetically pleasing shots will have sensible compositions that respect symmetry and certain ratios in the scene. Violating these principles leaves behind another trace of the cropping operation.

These traces are very subtle,
and the human eye often cannot detect them, which makes studying and characterizing them challenging. However, neural networks are excellent at identifying these patterns. Indeed, extensive effort goes into preventing neural networks from learning such shortcuts enabled by image crops \cite{doersch2015unsupervised, noroozi2017representation}.

In this paper, we flip this around and declare that these shortcuts are not bugs, but instead an opportunity to dissect and understand the subtle clues left behind from image cropping. Capitalizing on a large, high-quality collection of natural images, we train a convolutional neural network to predict the absolute spatial location of a patch within an image. This is only possible if there exist visual features that are \emph{not} spatially invariant. Our experiments analyze the types of features that this model learns, and we show that it is possible to detect traces of the cropping operation. We can also use the discovered artefacts, along with semantic information, to recover where the crop was positioned in the original sensor plane.

\begin{figure}[t]
    \centering
    \begin{subfigure}{\linewidth}
        \centering
        \includegraphics[width=\linewidth, trim={1cm 1cm 1cm 1cm}, clip]{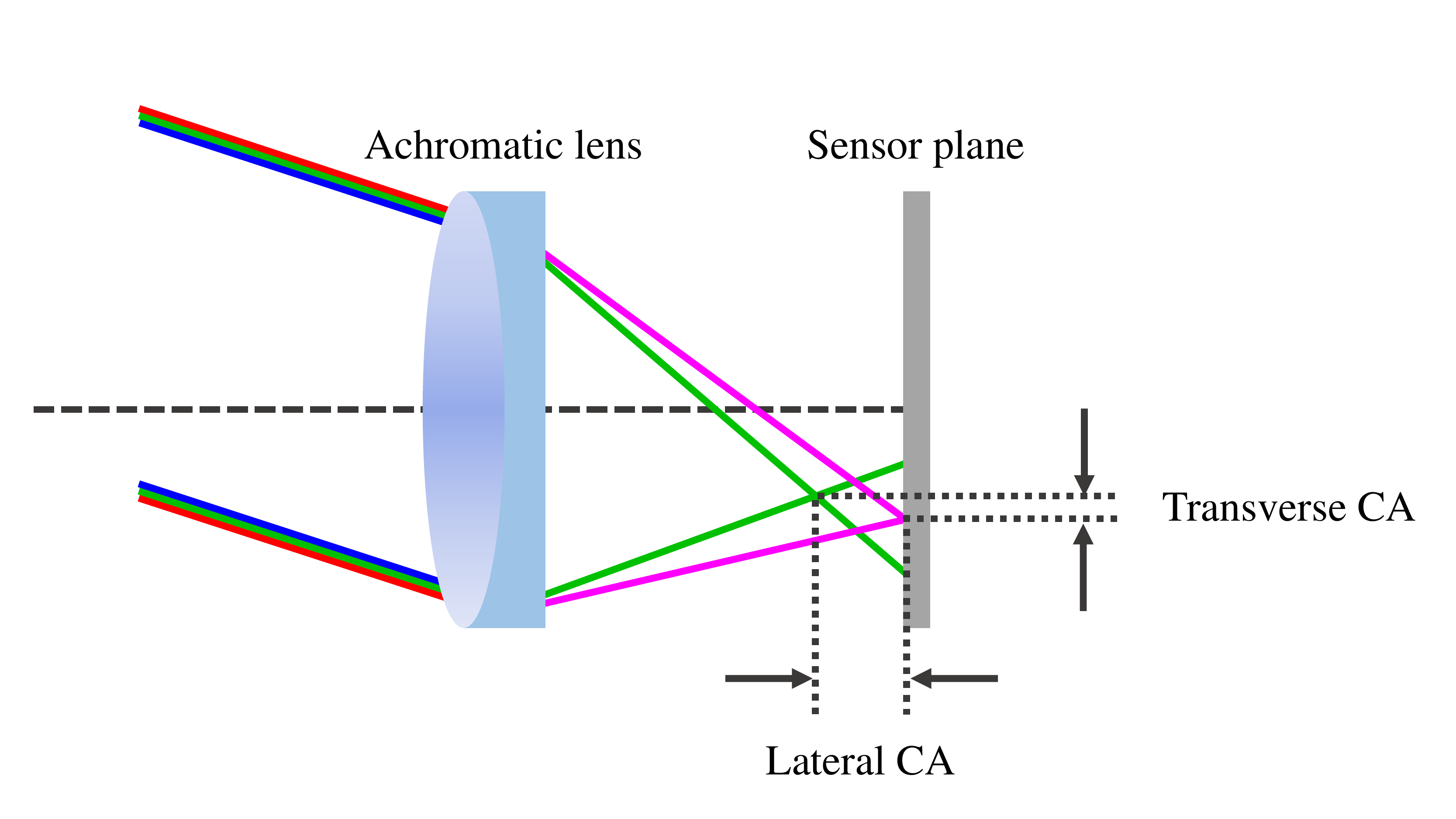}
        \caption{Lens with transverse and longitudinal chromatic aberration. In this illustration, the red and blue channels are aligned (hence the magenta rays), but green-colored light is magnified differently in addition to having a separate in-focus plane.}
        \label{fig:tca_lca}
    \end{subfigure}
    \par\medskip
    \begin{subfigure}{\linewidth}
        \centering
        \includegraphics[width=0.85\linewidth]{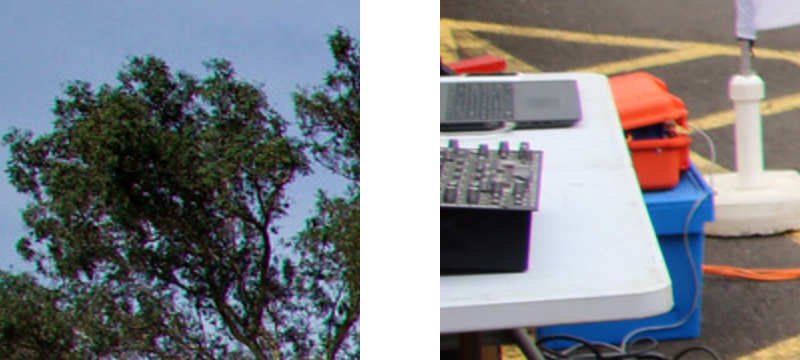}
        \caption{Close-up of two photos, revealing visible transverse chromatic aberration (TCA) artefacts.}
        \label{fig:chroma_zoom}
    \end{subfigure} 
    \caption{The origin behind, and examples of, chromatic aberration. \vspace{-0.2cm}}
    \label{fig:chroma}
\end{figure}

While the aim of this paper is to analyze the fundamental traces of image cropping in order to question conventional assumptions about translational invariance and the crucial role of data augmentation pervading the field, we believe our investigation could have a large practical impact as well. Historically, asking fundamental questions has spurred significant insight into core computer vision problems, such as invariances to scale \cite{burt1983laplacian}, asymmetries in time \cite{pickup2014seeing}, the speediness of videos \cite{benaim2020speednet}, and visual chirality \cite{lin2020visual}. For example, insight into image crops could enable detection of soft tampering operations, or spur developments to mitigate shortcut learning.

In Section 2, we briefly review related work in lens imperfections and self-supervised learning.
In Section 3, we describe the principles that govern our curation of an appropriate dataset.
In Section 4, we learn a representation for detecting absolute patch locations, and subsequently motivate the complete network architecture that incorporates both aggregated patch-wise and global context.
In Section 5, we analyze what our model has learned by measuring its performance under various controlled circumstances.
Finally, in Section 6, we visualize and interpret how the model addresses several interesting examples of image crops.

\section{Background and Related Work}

\textbf{Optical aberrations.} No imaging device is perfect, and every step in the imaging formation pipeline will leave traces behind onto the final picture. The origins of these signatures range from the physics of light in relation to the camera hardware, to the digital demosaicing and compression algorithms used to store and reconstruct the image. Lenses typically suffer from several aberrations, including chromatic aberration, vignetting, coma, and radial distortion \cite{kidger2001fundamental,beeson2007patterns,lin2018image,kashiwagi2019deep}. As shown in Figure \ref{fig:tca_lca}, chromatic aberration is manifested in two ways: \textit{transverse} (or \textit{lateral}) chromatic aberration (TCA) refers to the spatial discrepancies in focus points across color channels perpendicular to the optical axis, while \textit{longitudinal} chromatic aberration (LCA) refers to shifts in focus along the optical axis instead \cite{kang2007automatic,kashiwagi2019deep}. TCA gives rise to color channels that appear to be scaled slightly differently relative to each other, while LCA causes the distance between the focal surface and the lens to be frequency-dependent, such that the degree of blurring varies among color channels. Chromatic aberration can be leveraged to extract depth maps from defocus blur \cite{garcia2000chromatic,trouve2013passive,kashiwagi2019deep}, although the spatial sensitivity of these cues is often undesired \cite{doersch2015unsupervised,noroozi2016unsupervised,noroozi2017representation,nathan2018improvements}.
TCA is leveraged by \cite{yerushalmy2011digital} to measure the angle of an image region relative to the lens as a means to detect cropped images. We instead present a learning-based approach that discovers additional clues without the need for carefully tailored algorithms.

\begin{figure*}[b]
    \centering
    \includegraphics[width=\textwidth]{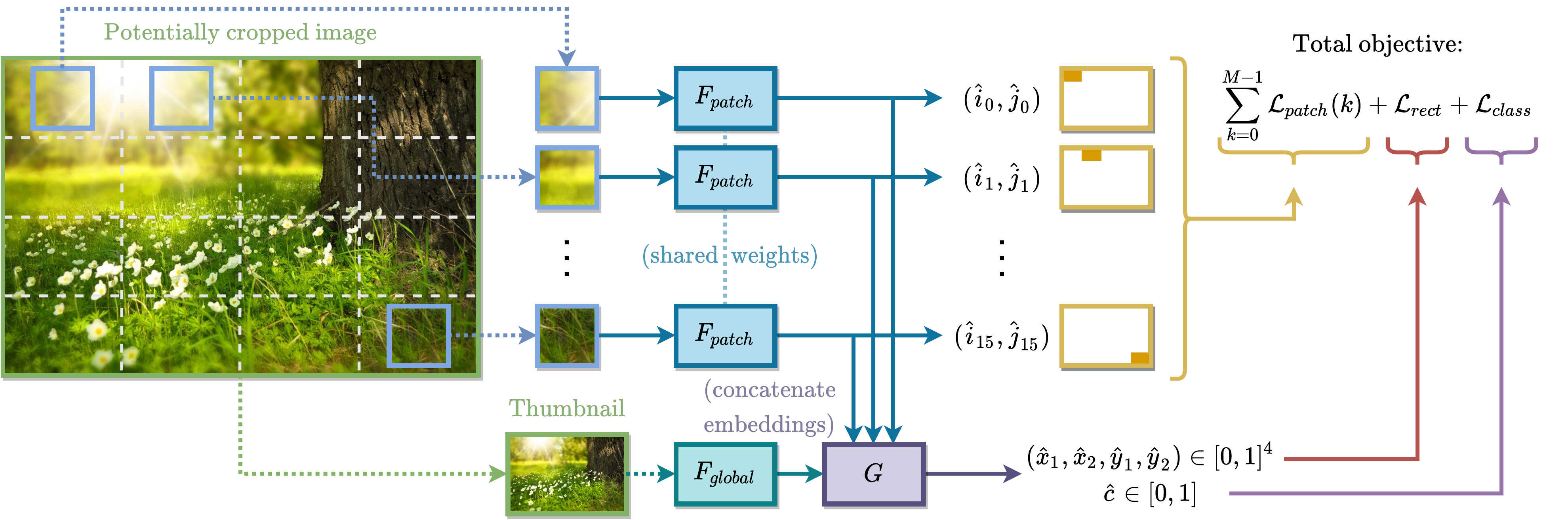}
    \caption{ 
    \textbf{Full architecture of our crop detection model.} We first extract $M=16$ patches from the centers of a regularly spaced grid within the source image, a priori not knowing whether it is cropped or not. The patch-based network $F_{patch}$ looks at each patch and classifies its absolute position into one out of 16 possibilities, whereby the estimation is mostly guided by low-level lens artefacts. The global image-based network, $F_{global}$ instead operates on the downscaled source image, and tends to pick up semantic signals, such as objects deviating from their canonical pose (\eg a face is cut in half). Since these two networks complement each other's strengths and weaknesses, we integrate their outputs into one pipeline via the multi-layer perceptron $G$. Note that $F_{patch}$ is supervised by all three loss terms, while $F_{global}$ only controls the crop rectangle $(\hat{x}_1, \hat{x}_2, \hat{y}_1, \hat{y}_2)$ and the final score $\hat{c}$.}
    \label{fig:arch}
\end{figure*}

\textbf{Patch localization.}
While one of the first major works in self-supervised representation learning focused on predicting the \textit{relative} location of two patches among eight possible configurations \cite{doersch2015unsupervised}, it was also discovered that the ability to perform \textit{absolute} localization seemed to arise out of chromatic aberration. For the best-performing 10\% of images, the mean Euclidean distance between the ground truth and predicted positions of single patches is 31\% lower than chance, and this gap narrowed to 13\% if every image was pre-processed to remove color information along the green-magenta axis. Although there are reasons to believe that modern network architectures might perform better,
these rather modest performance figures suggest a priori that the attempted task is a difficult one. Note that the learnability of absolute location is often regarded as a bug; treatments used in practice include random color channel dropping \cite{doersch2015unsupervised}, projection \cite{doersch2015unsupervised}, grayscale conversion \cite{noroozi2016unsupervised,noroozi2017representation}, jittering \cite{noroozi2016unsupervised}, and chroma blurring \cite{nathan2018improvements}.

\textbf{Visual crop detection.}
In the context of forensics, almost all existing research has centered around 'hard' tampering such as splicing and copy-move operations. We argue that some forms of 'soft' tampering, notably cropping, are also worth investigating.
While a few papers have addressed image crops \cite{yerushalmy2011digital, meng2013detecting, fanfani2020vision}, they are typically tailored toward specific types of pictures only. For example, both \cite{meng2013detecting} and \cite{fanfani2020vision} rely heavily on structured image content in the form of vanishing points and lines, which works only if many straight lines (\eg man-made buildings or rooms) are prominently visible in the scene.
Various previous works have also explored JPEG compression, and some have found that it may help reveal crops under specific circumstances, mostly by characterizing the regularity and alignment of blocking artefacts \cite{li2009passive, bruna2011crop, nguyen2013detecting}.
In contrast, our analysis focuses on camera pipeline artefacts and photography patterns that exist independently of digital post-processing algorithms.

\section{Dataset} \label{sec:dataset}

The natural clues for detecting crops are subtle, and we need to be careful to preserve them when constructing a dataset. Our underlying dataset has around $700,000$ high-resolution photos from Flickr, which were scraped during the fall of 2019. We impose several constraints on the training images, most importantly that they should not already have been cropped and that they must maintain a constant, fixed aspect ratio and resolution.
Appendix A describes this selection and collection process in detail.

We generate image crops by first defining
the \textit{crop rectangle} $(x_1, x_2, y_1, y_2)\in[0,1]^4$ as the relative boundaries of a cropped image within its original camera sensor plane, such that $(x_1, x_2, y_1, y_2)=(0,1,0,1)$ for unmodified images. We always maintain the aspect ratio and pick a random size factor $f$ uniformly in $[0.5, 0.9]$, representing the relative length in pixels of any of the four sides compared to the original photo: $f=x_2-x_1=y_2-y_1$.
Every crop touches an edge, such that the selected rectangle has a one in four chance of being positioned near either the top, right, bottom, or left border of the sensor plane.
After randomly cropping exactly half of all incoming photos, we give our model access to small image patches as well as global context.
We select square patches of size $96\times96$ (\ie around 5\% of the horizontal image dimension), which is sufficiently large to allow the network to get a good idea of the local texture profile, while also being small enough to ensure that neighbouring patches never overlap.
In addition, we downscale the whole image to a $224\times149$ thumbnail, such that it remains accessible to the model in terms of its receptive field and computational efficiency.\footnote{The reason we care about receptive field is because, even though high-resolution images are preferable when analyzing subtle lens artefacts, a ResNet-$L$ with $L\leq50$ has a receptive field of only $\leq483$ pixels \cite{araujo2019computing}, which pushes us to prefer \textit{lower} resolutions instead.}

\begin{figure*}[b]
    \centering
    \begin{subfigure}{\textwidth}
        \centering
        \includegraphics[width=\textwidth]{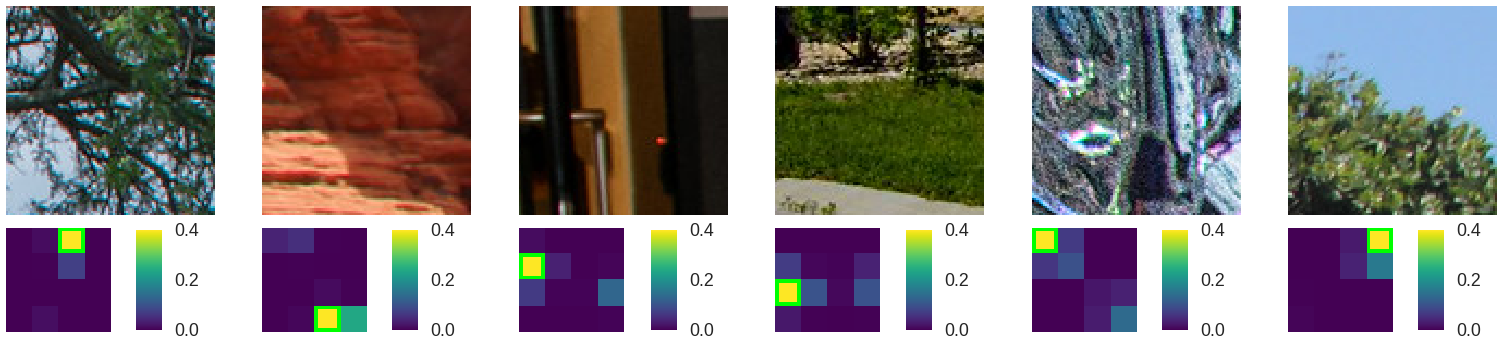}
        \caption{Selecting for \textbf{high confidence} yields samples biased toward highly textured content with many edges, often with visible chromatic aberration. The pretext model is typically more accurate in this case.}
        \label{fig:patch_high_score}
    \end{subfigure}
    \par\medskip
    \begin{subfigure}{\textwidth}
      \centering
      \includegraphics[width=\textwidth]{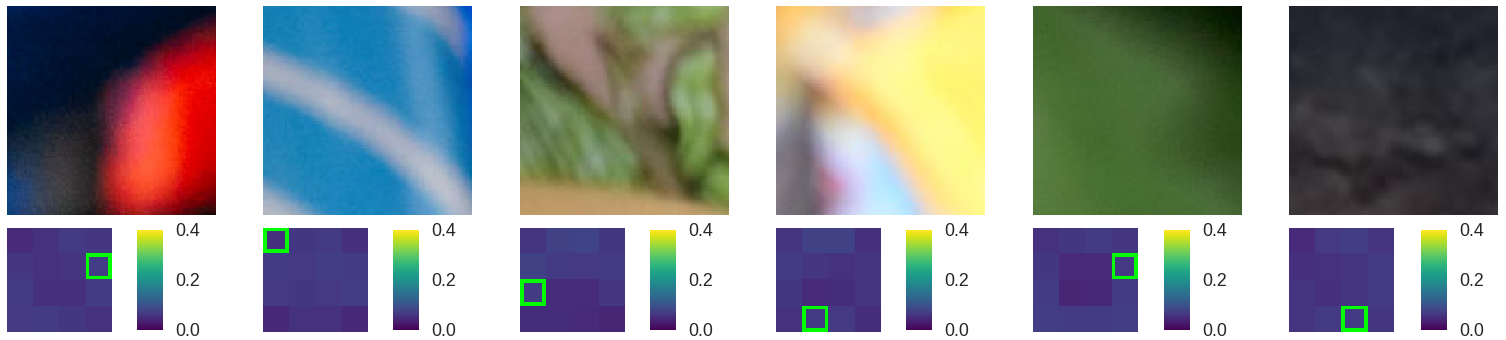}
      \caption{Selecting for \textbf{low confidence} yields blurry or smooth samples, where the lack of detail makes it difficult to expose physical imperfections of the lens. The pretext model tends to be inaccurate in this case.}
      \label{fig:patch_low_score}
    \end{subfigure} 
    \caption{
    \textbf{Absolute patch localization performance.}
    By leveraging classification, an uncertainty metric emerges for free. Here, we display examples where the pretext model $F_{patch}$ performs either exceptionally well or badly at recovering the patches' absolute position within the full image. The output probability distribution generated by the network is also plotted as a spatial heatmap
    ({\includegraphics[raise=28pt,valign=m,width=0.34cm]{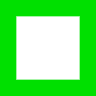}} = ground truth).}
    \label{fig:patch_high_low}
\end{figure*}

Interrelating contextual, semantic information to its spatial position within an image might turn out to be crucial for spotting crops. We therefore add coordinates as two extra channels to the thumbnail, similarly to \cite{liu2018intriguing}. Note that the model does not know a priori whether its input had been cropped or not. Lastly, several shortcut fuzzing procedures had to be used to ensure that the learned features are generalizable; see Appendix B for an extensive description.

\section{Approach}

We describe our methodology and the challenges associated with revealing whether and how a variably-sized single image has been cropped.
First, we construct a neural network that can trace image patches back to their original position relative to the center of the lens. Then, we use this novel network to expose and analyze possibly incomplete images using an end-to-end trained crop detection model, which also incorporates the global semantic context of an image in a way that can easily be visualized and understood. Figure \ref{fig:arch} illustrates our method.

\subsection{Predicting absolute patch location}
\label{sec:patch_loc}

One piece of the puzzle towards analyzing image crops is a neural network called $F_{patch}$, which discriminates the original position of a small image patch with respect to the center of the lens.
We frame this as a classification problem for practical purposes, and divide every image into a grid of $4\times4$ evenly sized cells, each of which represents a group of possible patch positions. Since this pretext task can be considered to be a form of self-supervised representation learning, with crop detection being the eventual downstream task, we call $F_{patch}$ the \textit{pretext model}.

But before embarking on an end-to-end crop detection journey that simply integrates this module into a larger system right from the beginning, it is worth asking the following questions: When exactly does absolute patch localization work well in the first place, and how could it help in distinguishing cropped images in an interpretable manner?
To this end, we trained $F_{patch}$ in isolation by discarding $F_{global}$ and forcing the network to decide based on information from patches only.
The 16-way classification loss term $\mathcal{L}_{patch}$ is responsible for pretext supervision, and is applied onto every patch individually.

Intriguing patterns emerge when discriminating between different levels of confidence in the predictions produced by $F_{patch}$. Although the accuracy of this localization network is not that high ({\raise.17ex\hbox{$\scriptstyle\sim$}}$21\%$ versus {\raise.17ex\hbox{$\scriptstyle\sim$}}$6\%$ for chance) due to the inherent difficulty of the task, Figure \ref{fig:patch_high_low} shows that it works quite well for some images, particularly those with a high degree of detail coupled with apparent lens artefacts. On the flip side, blurry photos taken with high-end cameras tend to make the model uncertain.
This observation suggests that chromatic aberration has strong predictive power for the original locations of patches within pictures. Hence, it is reasonable to expect that incorporating patch-wise, pixel-level cues into a deep learning-based crop detection framework will improve its capabilities.

\subsection{Architecture and objective}

Guided by the design considerations laid out so far, Figure \ref{fig:arch} shows our main model architecture.
$F_{patch}$ is a ResNet-18 \cite{he2016deep} that converts any patch into a length-64 embedding, which then gets converted by a single linear layer on top to a length-16 probability distribution describing the estimated location $(\hat{i}_k,\hat{j}_k)\in\{0\ldots 3\}^2$ of that patch. $F_{global}$ is a ResNet-34 \cite{he2016deep} that converts the downscaled global image into another length-64 embedding. Finally, $G$ is a 3-layer perceptron that accepts a 1088-dimensional concatenation of all previous embeddings, and produces 5 values describing (1) the crop rectangle $(\hat{x}_1, \hat{x}_2, \hat{y}_1, \hat{y}_2)\in[0,1]^4$, and (2) the actual probability $\hat{c}$ that the input image had been cropped.
By simultaneously processing and combining aggregated patch-wise information with global context, we allow the network to draw a complete picture of the input, revealing both low-level lens aberrations and high-level semantic cues.
The total, weighted loss function is as follows (with $M=16$):
\begin{align}
    \mathcal{L} &= \frac{\lambda_1}{M} \sum_{k=0}^{M-1} \mathcal{L}_{patch}(k) + \frac{\lambda_2}{4} \mathcal{L}_{rect} + \lambda_3 \mathcal{L}_{class}
\end{align}

Here, $\mathcal{L}_{patch}(k)$ is a 16-way cross-entropy classification loss between the predicted location distribution $\hat{\bm{l}}(k)$ of patch $k$ and its ground truth location $\bm{l}(k)$. For an uncropped image, $\bm{l}(k)=k$ and $(i_k,j_k)=(k\mod4,\floor{k/4})$, although this equality obviously does not necessarily hold for cropped images. Second, the loss term $\mathcal{L}_{rect}$ encourages the estimated crop rectangle to be near the ground truth in a mean squared error sense. Third, $\mathcal{L}_{class}$ is a binary cross-entropy classification loss that trains $\hat{c}$ to state whether or not the photo had been cropped. More formally:
\begin{align}
    \mathcal{L}_{patch}(k) &= \mathcal{L}_{CE}(\hat{\boldsymbol{l}}(k),\boldsymbol{l}(k)) \\
    \mathcal{L}_{rect} &= \big[ (\hat{x}_1-x_1)^2+(\hat{x}_2-x_2)^2 \nonumber \\
    &\hspace{0.75cm} + (\hat{y}_1-y_1)^2+(\hat{y}_2-y_2)^2 \big] \\
    \mathcal{L}_{class} &= \mathcal{L}_{BCE}(\hat{c},c)
\end{align}

Note that the intermediate outputs $(\hat{i}_k,\hat{j}_k)$ and $(\hat{x}_1, \hat{x}_2, \hat{y}_1, \hat{y}_2)$
exist mainly to encourage a degree of interpretability of the internal representation, rather than to improve the accuracy of the final score $\hat{c}$.
Specifically, the linear projection of $F_{patch}$ to $(\hat{i}_k,\hat{j}_k)$ should make the embedding more sensitive to positional information, thus helping the crop rectangle estimation.

\subsection{Training details}

In our experiments, all datasets are generated by cropping exactly 50\% of the photos with a random crop factor in $[0.5, 0.9]$. After that, we resize every example to a uniformly random width in $[1024, 2048]$ both during training and testing, such that the image size cannot have any predictive power. We train for up to 25 epochs using an Adam optimizer \cite{kingma2014adam}, with a learning rate that drops exponentially from $5\cdot10^{-3}$ to $1.5\cdot10^{-3}$ at respectively the first and last epoch. The weights of the loss terms are: $\lambda_1=2.4,\lambda_2=3,$ and $\lambda_3=1$.

\begin{figure*}
    \centering
    \begin{subfigure}{\textwidth}
        \centering
        \includegraphics[width=\textwidth]{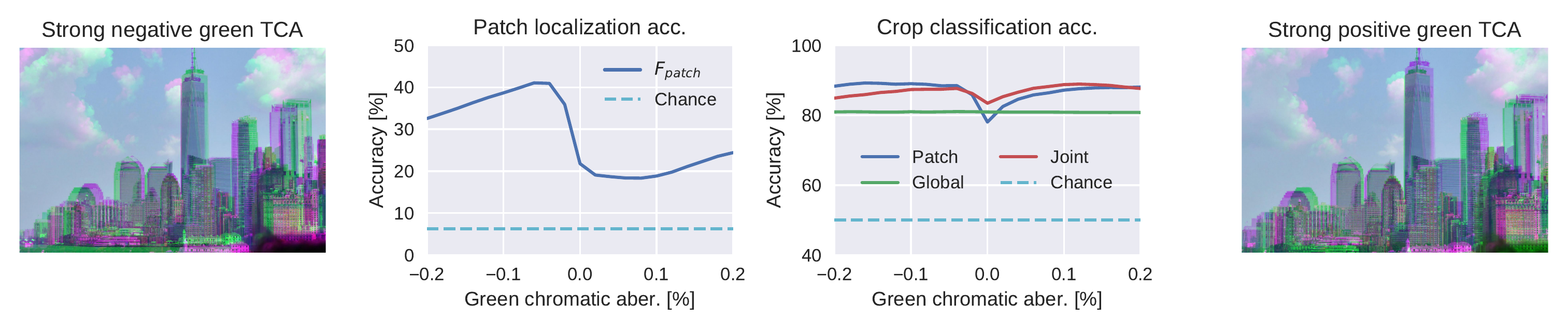}
        \caption{\textbf{Green transverse chromatic aberration} in the negative (inward) direction considerably boosts performance for patch localization, although asymmetry is key for crop detection. The \textit{global} model remains unaffected since it is unlikely to be able to see the artefacts. (We show examples with excessive distortion for illustration; the range used in practice is much more modest.)}
        \label{fig:sweep_gca}
    \end{subfigure}
    \par\medskip
    \begin{subfigure}{\textwidth}
      \centering
      \includegraphics[width=\textwidth]{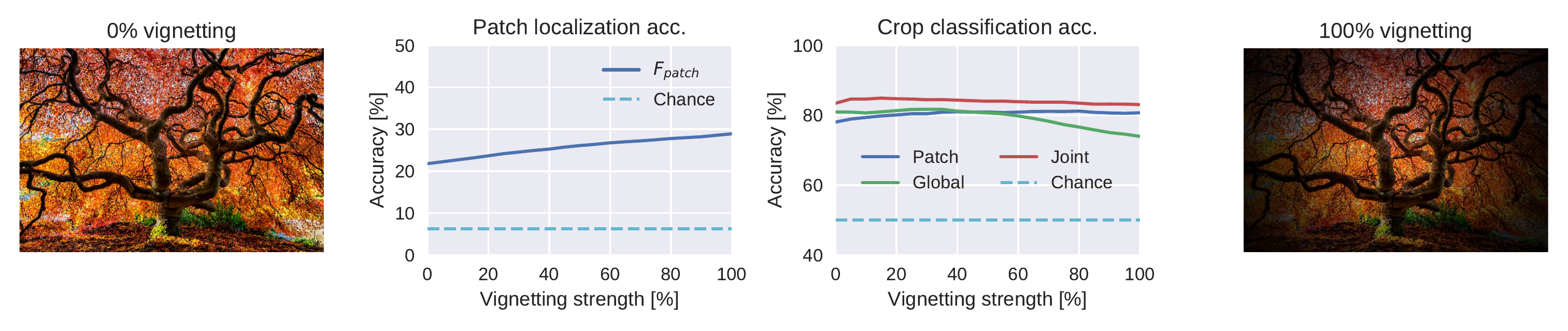}
      \caption{\textbf{Vignetting} also contributes positively to the pretext model’s accuracy. Interestingly, the crop detection performance initially increases but then drops slightly for strong vignetting, presumably because the distorted images are moving out-of-distribution.}
      \label{fig:sweep_vign}
    \end{subfigure}
    \caption{\textbf{Breakdown of image attributes that contribute to features relevant for crop detection.} In these experiments, we manually exaggerate two characteristics of the lens on 3,500 photos of the test set, and subsequently measure the resulting shift in performance.
    }
    \label{fig:sweep}
\end{figure*}

\section{Analysis and Clues} \label{sec:results}

We quantitatively investigate the model in order to dissect and characterize visual crops.
We are interested in conducting a careful analysis of what factors the network might be looking at within every image. 
For ablation study purposes, we distinguish three variants of our model:
\begin{itemize}
    \item \textbf{Joint} is the complete patch- and global-based model from Figure \ref{fig:arch} central to this work;
    \item \textbf{Global} is a naive classifier that just operates on the thumbnail, \ie the whole input downscaled to $224\times149$, using $F_{global}$;
    \item \textbf{Patch} only sees $16$ small patches extracted from consistent positions within the image, using $F_{patch}$.
\end{itemize}

We classify the information that a model uses as evidence for its decision into two broad categories: \textbf{(1) characteristics of the camera or lens system}, and \textbf{(2) object priors}. While (1) is largely invariant of semantic image content, (2) could mean that the network has learned to leverage certain rules in photography, \eg the sky is usually on top, and a person's face is usually centered.

To gain insight into what exactly our model has discovered, we first investigate the network's response to several known lens characteristics by artificially inflating their corresponding optical aberrations on the test set, and computing the resulting performance metrics. Next, we measure the changes in accuracy when the model is applied on datasets that were crafted specifically as to have divergent distributions over object semantics and image structure. We expect both lens flaws and photographic conventions to play different but interesting roles in our model.

A discussion of chromatic aberration expressed along the green channel, vignetting, and photography patterns follows; see Appendix C for the effect of color saturation, radial lens distortion, and chromatic aberration of the red and blue channels. Note that all discussed image modifications are applied \emph{prior} to cropping, as a means of simulating a real lens that exhibits certain controllable defects.

\subsection{Effect of chromatic aberration} \label{sec:chroma}

A common lens correction to counter the frequency-dependence of the refractive index of glass is to use a so-called \textit{achromatic doublet}. This modification ensures that the light rays of two different frequencies, such as the red and blue color channels, are aligned \cite{kidger2001fundamental}. Because the remaining green channel still undergoes TCA and will therefore be slightly downscaled around the optical center, this artefact is often visible as green or purple fringes near edges and other regions with contrast or texture \cite{brewster1833treatise}. Figure \ref{fig:chroma_zoom} depicts real examples of what chromatic aberration looks like. Note that the optical center around which radial magnification occurs does not necessarily coincide with the image center due to the complexity of multi-lens systems \cite{willson1994center}, although both points have been found to be very close in practice \cite{kang2007automatic}. Furthermore, chromatic aberration can vary strongly from device to device, and is not even present in all camera systems. Many high-end, modern lenses and/or post-processing algorithms tend to accurately correct for them, to the point that it becomes virtually imperceptible.

Nonetheless, our model still finds this spectral discrepancy in focus points to be a distinctive feature of crops and patch positions:
Figure \ref{fig:sweep_gca} (left plot) demonstrates that artificially downscaling the green channel significantly improves the pretext model's performance. This is because the angle and magnitude of texture shifts across color channels can give away the location of a patch relative to the center of the lens. Consequently, the downstream task of crop detection (right plot) becomes easier when TCA is introduced in either direction.
Horizontally mirrored plots were obtained upon examining the red and blue channels, confirming that the green channel suffers an inward deviation most commonly of all in our dataset. It turns out that the optimal configuration from the perspective of $F_{patch}$ is to add a little distortion, but not too much --- otherwise we risk hurting the realism of the test set.

\begin{table}
    \centering
        \begin{tabular}{l r r r r}
            \toprule
    \textbf{Dataset} & \textbf{Joint} & \textbf{Global} & \textbf{Patch} & \textbf{Human} \\ \midrule
    Flickr & 86\% & 79\% & 77\% & 67\% \\
    Flickr (no humans) & 81\% & 75\% & 73\% & - \\ \midrule
    Upright & 80\% & 72\% & 76\% & - \\
    Tilted & 71\% & 58\% & 70\% & - \\
    Vanish & 82\% & 75\% & 79\% & - \\
    Texture & 66\% & 54\% & 67\% & - \\
    Smooth & 50\% & 51\% & 55\% & - \\
            \bottomrule
        \end{tabular}
        \caption{\textbf{Accuracy comparison between three different crop detection models on various datasets.} All models are trained on Flickr, and appear to have discovered common rules in photography to varying degrees. \\[-0.24in]
        }
        \label{tab:crop_cmp}
    \end{table}

\begin{figure*}
    \centering
    \includegraphics[width=\textwidth]{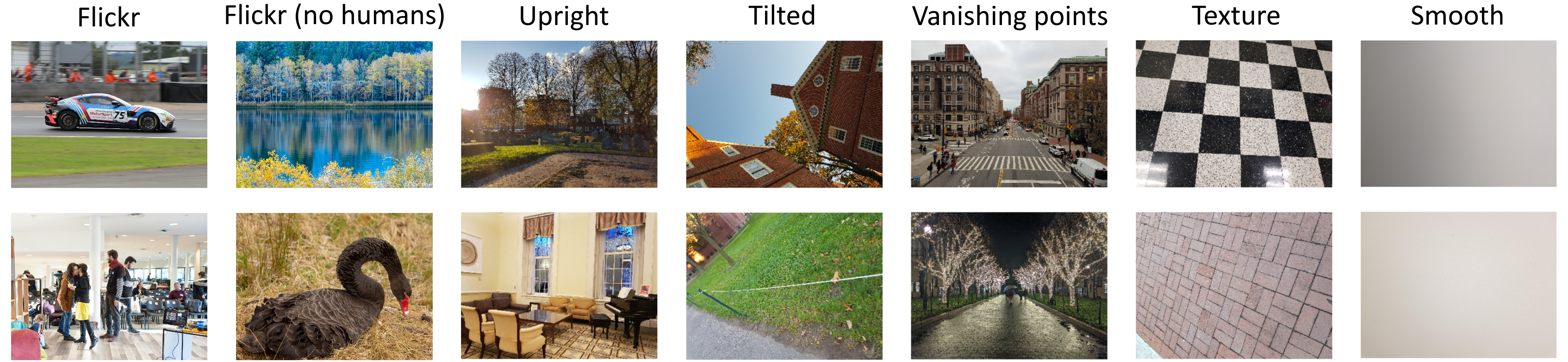}
    \caption{\textbf{Representative examples of the seven test sets.} The first two are variants of Flickr, one unfiltered and one without humans or faces, and the remaining five are custom photo collections we intend to measure various other kinds of photographic patterns or biases with. These were taken in New York, Boston, and SF Bay Area, and every category contains between 15 and 127 pictures. \\[-0.24in]
    }
    \label{fig:datasets_images}
\end{figure*}

\subsection{Effect of vignetting}

A typical imperfection of multi-lens systems is the radial brightness fall-off as we move away from the center of the image, seen in Figure \ref{fig:sweep_vign}. Vignetting can arise due to mechanical and natural reasons \cite{lopez2015revisiting}, but its dependence on the position within a photo is the most important aspect in this context. We simulate vignetting by multiplying every pixel value with $\frac{1}{g(r)}$, where:
\begin{align}
    g(r) &= 1+ar^2+br^4+cr^6 \label{eq:vign} \\
    (a,b,c) &= (2.0625, 8.75, 0.0313)
\end{align}
$g(r)$ is a sixth-grade polynomial gain function, the parameters $a,b,c$ are assigned typical values taken from \cite{lopez2015revisiting}, and $r$ represents the radius from the image center with $r=1$ at every corner. The degree of vignetting is smoothly varied by simply interpolating every pixel between its original (0\%) and fully modified (100\%) state.

Figure \ref{fig:sweep_vign} shows that enhanced vignetting has a positive impact on absolute patch localization ability, but this does not appear to translate into noticeably better crop detection accuracy. While the gradient direction of the brightness across a patch is a clear indicator of the angle that it makes with respect to the optical center of the image, modern cameras appear to correct for vignetting well enough such that the lack of realism of the perturbed images hurts $F_{global}$'s performance more so than it helps.

\subsection{Effect of photography patterns and perspective}
\label{sec:photobias}

The desire to capture meaningful content implies that not all images are created equal. Interesting objects, persons, or animals will often intentionally be centered within a photo, and cameras are generally oriented upright when taking pictures. Some conventions, \eg grass is usually at the bottom, are confounded to some extent by the random rotations during training, although there remain many facts to be learned as to what constitutes an appealing or sensible photograph. One clear example of these so-called \textit{photography patterns} in the context of our model is that when a person's face that is cut in half, this might reveal that the image had been cropped. This is because, intuitively speaking, it does not conform to how photographers typically organize their visual environment and constituents of the scene. 

The structure of the world around us not only provides high-level knowledge on where and how objects typically exist within pictures, but also gives rise to perspective cues, for example the angle that horizontal lines make with vertical lines upon projection of a 3D scene onto the 2D sensor, coupled with the apparent normal vector of a wall or other surface.
Measuring the exact extent to which all of these aspects play a role is difficult, as no suitable dataset exists. The ideal baseline would consist of photos without any adherence to photography rules whatsoever, taken in uniformly random orientations at arbitrary, mostly uninteresting locations around the world.

We constructed and categorized a small-scale collection of such photos ourselves, using the Samsung Galaxy S8 and Google Pixel 4 smartphones, spanning the 5 right-most columns in Figure \ref{fig:datasets_images}.
Columns 3 and 5 depict photos that are taken with the camera in an upright, biased orientation. Column 5 specifically encompasses vanishing line-heavy content, where perspective clues may provide clear pointers. Columns 4, 6, and 7 contain pictures that are unlikely to be taken by a normal photographer, but whose purpose is instead to measure the response of our system on photos with compositions that make less sense.

\begin{figure*}[!t]
    \centering
    \includegraphics[width=\textwidth]{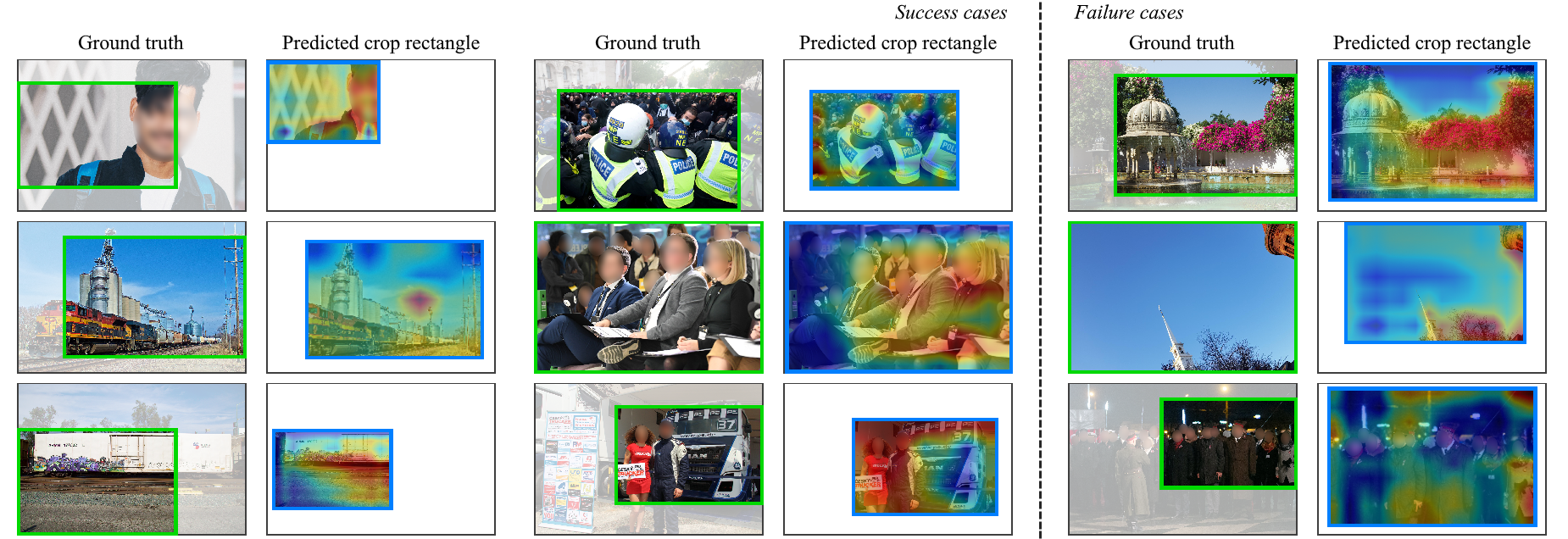}
    \caption{\textbf{Qualitative examples and interpretation of our crop detection system.} High-level cues such as persons and faces appear to considerably affect the model's decisions. Note that images don't always \textit{look} cropped, but in that case, patches can act as the giveaway whenever they express lens artefacts. Regardless, certain scene compositions are more difficult to get right, such as in the failure cases shown on the right. (Faced blurred here for privacy protection.)
    \\[-0.24in]
    }
    \label{fig:analysis_gradcam}
\end{figure*}

Quantitative results are shown in Table \ref{tab:crop_cmp}.
On the Flickr test set, the crop classification accuracy is 79\% for the thumbnail-based model, 77\% for the patch-based model, and 86\% for the joint model.
For comparison purposes, we also asked 16 people to classify 100 random Flickr photos into whether they look cropped or not, resulting in a human accuracy of 67\%. This demonstrates that integrating information across multiple scales results in a better model than a network that only sees either patches or thumbnails independently, in addition to having a significant performance margin over humans.

Our measurements also indicate that the model tends to consistently perform better on sensible, upright photos. Analogous to what makes many datasets curated \cite{caron2019unsupervised, selvaraju2020casting}, Flickr in particular seems to exhibit a high degree of photographic conventions involving people, so we also tested a manually filtered subset of 100 photos that do not contain humans or faces, resulting in a modest drop in accuracy.
Interestingly, the patch-based network comes very close to the joint network on \textit{tilted} and \textit{texture}, suggesting that global context can sometimes confuse the model if the photo is taken in an abnormal way.
Fully smooth, white-wall images appear to be even more out-of-distribution.
However, most natural imagery predominantly contains canonical and appealing arrangements, where our model displays a stronger ability to distinguish crops.

\section{Visualizing Image Crops}

\begin{figure}
    \centering
    \includegraphics[width=\linewidth]{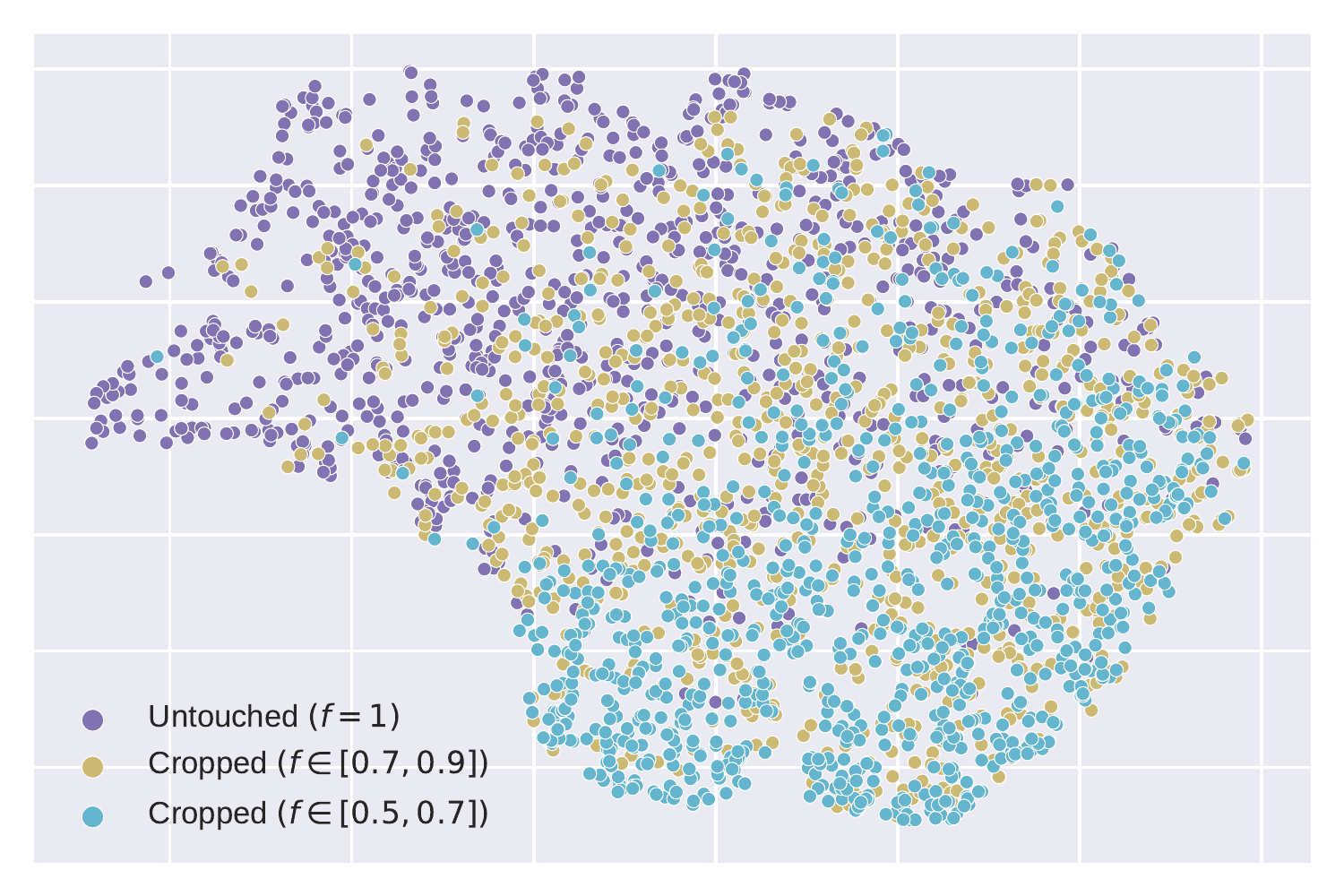}
    \caption{\textbf{Dimensionality-reduced embeddings generated by $F_{global}$ on Flickr.} Here, the size factor $f$ stands for the fraction of one cropped image dimension relative to the original photo. The model is clearly able to separate untampered from strongly cropped images, although lightly cropped images can land almost anywhere across the spectrum as the semantic signals might be less pronounced and/or less frequently present. \\[-0.24in]
    }
    \label{fig:embeddings}
\end{figure}

In order to depict the changing visual distribution as images are cropped to an increasingly stronger extent, we look at the output embeddings produced by the thumbnail network $F_{global}$. In Figure \ref{fig:embeddings}, we first apply Principal Component Analysis (PCA) to transform the data points from 64 to 24 dimensions, and subsequently apply t-SNE \cite{maaten2008visualizing} to further reduce the dimensionality from 24 to 2.

As discussed in the previous sections, there could be many reasons as to why the model predicts that a certain photo appears or does not appear to be cropped.
However, to explain results obtained from any given single input, we can also apply the Grad-CAM technique \cite{selvaraju2017grad} onto the global image.
This procedure allows us to construct a heatmap that attributes decisions made by $F_{global}$ and $G$ back to the input regions that contributed to them.

Figure \ref{fig:analysis_gradcam} showcases a few examples, where we crop untouched images by the green ground truth rectangle and subsequently feed them into the network to visualize its prediction.
The model is often able to \textit{uncrop} the image, using semantic and/or patch-based clues, and produce a reasonable estimate of which spatial regions are missing (if any). For example, the top left image clearly violates routine principles in photography.
The top or bottom images are a little harder to judge by the same measure, though we can still recover the crop frame thanks to the absolute patch localization functionality.

\section{Discussion}

We found that image regions contain information about their spatial position relative to the lens, refining established assumptions about translational invariance \cite{lecun2015deep}.
Our network has automatically discovered various relevant clues, ranging from subtle lens flaws to photographic priors.
These features are likely to be acquired to some extent by many self-supervised representation learning methods, such as contrastive learning, where cropping is an important form of data augmentation \cite{chen2020simple, selvaraju2020casting}.
Although they are often treated as a bug, there are also compelling cases where the clues could prove to be useful.
For example, we believe that our crop detection and analysis framework has implications for revealing misleading photojournalism.
We also hope that our work inspires further research into how the traces left behind by image cropping, and the altered visual distributions that it gives rise to, can be leveraged in other interesting ways.

{\small
\textbf{Acknowledgments:} We thank Dídac Surís, Chengzhi Mao, Mia Chiquier, and Abby Lu for helpful feedback. This research is based on work partially supported by NSF CRII Award \#1850069, the DARPA SAIL-ON program under PTE Federal Award No.\ W911NF2020009, and an Amazon Research Gift. We thank NVIDIA for GPU donations. Part of this work was performed while being the recipient of a Belgian American Educational Foundation Fellowship to BVH. The views and conclusions contained herein are those of the authors and should not be interpreted as necessarily representing the official policies, either expressed or implied, of the U.S. Government.
}

{\small
\bibliographystyle{ieee_fullname}
\bibliography{_bib}

\begin{thebibliography}{10}\itemsep=-1pt

\bibitem{opencv}
Opencv: Geometric image transformations.

\bibitem{mit_course_cv}
Alexander Amini and Ava Soleimany.
\newblock Mit 6.s191: Introduction to deep learning, spring 2020.

\bibitem{araujo2019computing}
André Araujo, Wade Norris, and Jack Sim.
\newblock Computing receptive fields of convolutional neural networks.
\newblock {\em Distill}, 2019.
\newblock https://distill.pub/2019/computing-receptive-fields.

\bibitem{barbu2019objectnet}
Andrei Barbu, David Mayo, Julian Alverio, William Luo, Christopher Wang, Dan
  Gutfreund, Josh Tenenbaum, and Boris Katz.
\newblock Objectnet: A large-scale bias-controlled dataset for pushing the
  limits of object recognition models.
\newblock In {\em Advances in Neural Information Processing Systems}, pages
  9453--9463, 2019.

\bibitem{beeson2007patterns}
Steven Beeson and James~W Mayer.
\newblock {\em Patterns of light: chasing the spectrum from Aristotle to LEDs}.
\newblock Springer Science \& Business Media, 2007.

\bibitem{benaim2020speednet}
Sagie Benaim, Ariel Ephrat, Oran Lang, Inbar Mosseri, William~T Freeman,
  Michael Rubinstein, Michal Irani, and Tali Dekel.
\newblock Speednet: Learning the speediness in videos.
\newblock In {\em Proceedings of the IEEE/CVF Conference on Computer Vision and
  Pattern Recognition}, pages 9922--9931, 2020.

\bibitem{brewster1833treatise}
David Brewster and Alexander~Dallas Bache.
\newblock {\em A Treatise on Optics...: First American Edition, with an
  Appendix, Containing an Elementary View of the Application of Analysis to
  Reflexion and Refraction}.
\newblock Carey, Lea, \& Blanchard, 1833.

\bibitem{bruna2011crop}
AR Bruna, Giuseppe Messina, and Sebastiano Battiato.
\newblock Crop detection through blocking artefacts analysis.
\newblock In {\em International Conference on Image Analysis and Processing},
  pages 650--659. Springer, 2011.

\bibitem{burt1983laplacian}
Peter Burt and Edward Adelson.
\newblock The laplacian pyramid as a compact image code.
\newblock {\em IEEE Transactions on communications}, 31(4):532--540, 1983.

\bibitem{canales_2020}
Katie Canales.
\newblock Twitter is making changes to its photo software after people online
  found it was automatically cropping out black faces and focusing on white
  ones, Oct 2020.

\bibitem{caron2019unsupervised}
Mathilde Caron, Piotr Bojanowski, Julien Mairal, and Armand Joulin.
\newblock Unsupervised pre-training of image features on non-curated data.
\newblock In {\em Proceedings of the IEEE/CVF International Conference on
  Computer Vision}, pages 2959--2968, 2019.

\bibitem{chen2016automatic}
Jiansheng Chen, Gaocheng Bai, Shaoheng Liang, and Zhengqin Li.
\newblock Automatic image cropping: A computational complexity study.
\newblock In {\em Proceedings of the IEEE Conference on Computer Vision and
  Pattern Recognition}, pages 507--515, 2016.

\bibitem{chen2020simple}
Ting Chen, Simon Kornblith, Mohammad Norouzi, and Geoffrey Hinton.
\newblock A simple framework for contrastive learning of visual
  representations.
\newblock {\em arXiv preprint arXiv:2002.05709}, 2020.

\bibitem{sherling_2020}
London~Broadcasting Company.
\newblock Bbc criticised for cropping out weapon in black lives matter protest
  photo, Jun 2020.

\bibitem{doersch2015unsupervised}
Carl Doersch, Abhinav Gupta, and Alexei~A Efros.
\newblock Unsupervised visual representation learning by context prediction.
\newblock In {\em Proceedings of the IEEE international conference on computer
  vision}, pages 1422--1430, 2015.

\bibitem{esfandiari_martin_2020}
Sahar Esfandiari and Will Martin.
\newblock Greta thunberg slammed the associated press for cropping a black
  activist out of a photo of her at davos, Jan 2020.

\bibitem{fanfani2020vision}
Marco Fanfani, Massimo Iuliani, Fabio Bellavia, Carlo Colombo, and Alessandro
  Piva.
\newblock A vision-based fully automated approach to robust image cropping
  detection.
\newblock {\em Signal Processing: Image Communication}, 80:115629, 2020.

\bibitem{Franz2006}
Alex Franz and Thorsten Brants.
\newblock All our n-gram are belong to you, Aug 2006.

\bibitem{garcia2000chromatic}
Josep Garcia, Juan~Maria Sanchez, Xavier Orriols, and Xavier Binefa.
\newblock Chromatic aberration and depth extraction.
\newblock In {\em Proceedings 15th International Conference on Pattern
  Recognition. ICPR-2000}, volume~1, pages 762--765. IEEE, 2000.

\bibitem{he2020momentum}
Kaiming He, Haoqi Fan, Yuxin Wu, Saining Xie, and Ross Girshick.
\newblock Momentum contrast for unsupervised visual representation learning.
\newblock In {\em Proceedings of the IEEE/CVF Conference on Computer Vision and
  Pattern Recognition}, pages 9729--9738, 2020.

\bibitem{he2016deep}
Kaiming He, Xiangyu Zhang, Shaoqing Ren, and Jian Sun.
\newblock Deep residual learning for image recognition.
\newblock In {\em Proceedings of the IEEE conference on computer vision and
  pattern recognition}, pages 770--778, 2016.

\bibitem{hendrycks2019natural}
Dan Hendrycks, Kevin Zhao, Steven Basart, Jacob Steinhardt, and Dawn Song.
\newblock Natural adversarial examples.
\newblock {\em arXiv preprint arXiv:1907.07174}, 2019.

\bibitem{kang2007automatic}
Sing~Bing Kang.
\newblock Automatic removal of chromatic aberration from a single image.
\newblock In {\em 2007 IEEE Conference on Computer Vision and Pattern
  Recognition}, pages 1--8. IEEE, 2007.

\bibitem{kashiwagi2019deep}
Masako Kashiwagi, Nao Mishima, Tatsuo Kozakaya, and Shinsaku Hiura.
\newblock Deep depth from aberration map.
\newblock In {\em Proceedings of the IEEE International Conference on Computer
  Vision}, pages 4070--4079, 2019.

\bibitem{Kaufman2019}
Josh Kaufman.
\newblock Github: first20hours/google-10000-english, Aug 2019.

\bibitem{kidger2001fundamental}
Michael~J Kidger.
\newblock Fundamental optical design.
\newblock In {\em Fundamental optical design}. SPIE Bellingham, 2001.

\bibitem{kingma2014adam}
Diederik~P Kingma and Jimmy Ba.
\newblock Adam: A method for stochastic optimization.
\newblock {\em arXiv preprint arXiv:1412.6980}, 2014.

\bibitem{krizhevsky2017imagenet}
Alex Krizhevsky, Ilya Sutskever, and Geoffrey~E Hinton.
\newblock Imagenet classification with deep convolutional neural networks.
\newblock {\em Communications of the ACM}, 60(6):84--90, 2017.

\bibitem{kuznetsova2018open}
Alina Kuznetsova, Hassan Rom, Neil Alldrin, Jasper Uijlings, Ivan Krasin, Jordi
  Pont-Tuset, Shahab Kamali, Stefan Popov, Matteo Malloci, Tom Duerig, et~al.
\newblock The open images dataset v4: Unified image classification, object
  detection, and visual relationship detection at scale.
\newblock {\em arXiv preprint arXiv:1811.00982}, 2018.

\bibitem{lecun2015deep}
Yann LeCun, Yoshua Bengio, and Geoffrey Hinton.
\newblock Deep learning.
\newblock {\em nature}, 521(7553):436--444, 2015.

\bibitem{stanford_course_cv}
Fei-Fei Li, Ranjay Krishna, and Danfei Xu.
\newblock Stanford cs231n: Convolutional neural networks for visual
  recognition, spring 2020.

\bibitem{li2009passive}
Weihai Li, Yuan Yuan, and Nenghai Yu.
\newblock Passive detection of doctored jpeg image via block artifact grid
  extraction.
\newblock {\em Signal Processing}, 89(9):1821--1829, 2009.

\bibitem{lin2018image}
Xufeng Lin and Chang-Tsun Li.
\newblock Image provenance inference through content-based device fingerprint
  analysis.
\newblock In {\em Information Security: Foundations, Technologies and
  Applications}, pages 279--310. IET, 2018.

\bibitem{lin2020visual}
Zhiqiu Lin, Jin Sun, Abe Davis, and Noah Snavely.
\newblock Visual chirality.
\newblock In {\em Proceedings of the IEEE/CVF Conference on Computer Vision and
  Pattern Recognition}, pages 12295--12303, 2020.

\bibitem{liu2018intriguing}
Rosanne Liu, Joel Lehman, Piero Molino, Felipe~Petroski Such, Eric Frank, Alex
  Sergeev, and Jason Yosinski.
\newblock An intriguing failing of convolutional neural networks and the
  coordconv solution.
\newblock In {\em Advances in Neural Information Processing Systems}, pages
  9605--9616, 2018.

\bibitem{lopez2015revisiting}
Laura Lopez-Fuentes, Gabriel Oliver, and Sebastia Massanet.
\newblock Revisiting image vignetting correction by constrained minimization of
  log-intensity entropy.
\newblock In {\em International Work-Conference on Artificial Neural Networks},
  pages 450--463. Springer, 2015.

\bibitem{maaten2008visualizing}
Laurens van~der Maaten and Geoffrey Hinton.
\newblock Visualizing data using t-sne.
\newblock {\em Journal of machine learning research}, 9(Nov):2579--2605, 2008.

\bibitem{marra2020full}
Francesco Marra, Diego Gragnaniello, Luisa Verdoliva, and Giovanni Poggi.
\newblock A full-image full-resolution end-to-end-trainable cnn framework for
  image forgery detection.
\newblock {\em IEEE Access}, 8:133488--133502, 2020.

\bibitem{meng2013detecting}
Xianzhe Meng, Shaozhang Niu, Ru Yan, and Yezhou Li.
\newblock Detecting photographic cropping based on vanishing points.
\newblock {\em Chinese Journal of Electronics}, 22(2):369--372, 2013.

\bibitem{nathan2018improvements}
T Nathan~Mundhenk, Daniel Ho, and Barry~Y Chen.
\newblock Improvements to context based self-supervised learning.
\newblock In {\em Proceedings of the IEEE Conference on Computer Vision and
  Pattern Recognition}, pages 9339--9348, 2018.

\bibitem{nguyen2013detecting}
Hieu~Cuong Nguyen and Stefan Katzenbeisser.
\newblock Detecting resized double jpeg compressed images--using support vector
  machine.
\newblock In {\em IFIP International Conference on Communications and
  Multimedia Security}, pages 113--122. Springer, 2013.

\bibitem{noroozi2016unsupervised}
Mehdi Noroozi and Paolo Favaro.
\newblock Unsupervised learning of visual representations by solving jigsaw
  puzzles.
\newblock In {\em European Conference on Computer Vision}, pages 69--84.
  Springer, 2016.

\bibitem{noroozi2017representation}
Mehdi Noroozi, Hamed Pirsiavash, and Paolo Favaro.
\newblock Representation learning by learning to count.
\newblock In {\em Proceedings of the IEEE International Conference on Computer
  Vision}, pages 5898--5906, 2017.

\bibitem{oliva2001modeling}
Aude Oliva and Antonio Torralba.
\newblock Modeling the shape of the scene: A holistic representation of the
  spatial envelope.
\newblock {\em International journal of computer vision}, 42(3):145--175, 2001.

\bibitem{pathak2016context}
Deepak Pathak, Philipp Krahenbuhl, Jeff Donahue, Trevor Darrell, and Alexei~A
  Efros.
\newblock Context encoders: Feature learning by inpainting.
\newblock In {\em Proceedings of the IEEE conference on computer vision and
  pattern recognition}, pages 2536--2544, 2016.

\bibitem{pickup2014seeing}
Lyndsey~C Pickup, Zheng Pan, Donglai Wei, YiChang Shih, Changshui Zhang, Andrew
  Zisserman, Bernhard Scholkopf, and William~T Freeman.
\newblock Seeing the arrow of time.
\newblock In {\em Proceedings of the IEEE Conference on Computer Vision and
  Pattern Recognition}, pages 2035--2042, 2014.

\bibitem{samii2015data}
A Samii, R M{\v{e}}ch, and Zhe Lin.
\newblock Data-driven automatic cropping using semantic composition search.
\newblock In {\em Computer graphics forum}, volume~34, pages 141--151. Wiley
  Online Library, 2015.

\bibitem{selvaraju2017grad}
Ramprasaath~R Selvaraju, Michael Cogswell, Abhishek Das, Ramakrishna Vedantam,
  Devi Parikh, and Dhruv Batra.
\newblock Grad-cam: Visual explanations from deep networks via gradient-based
  localization.
\newblock In {\em Proceedings of the IEEE international conference on computer
  vision}, pages 618--626, 2017.

\bibitem{selvaraju2020casting}
Ramprasaath~R Selvaraju, Karan Desai, Justin Johnson, and Nikhil Naik.
\newblock Casting your model: Learning to localize improves self-supervised
  representations.
\newblock {\em arXiv preprint arXiv:2012.04630}, 2020.

\bibitem{szegedy2015going}
Christian Szegedy, Wei Liu, Yangqing Jia, Pierre Sermanet, Scott Reed, Dragomir
  Anguelov, Dumitru Erhan, Vincent Vanhoucke, and Andrew Rabinovich.
\newblock Going deeper with convolutions.
\newblock In {\em Proceedings of the IEEE conference on computer vision and
  pattern recognition}, pages 1--9, 2015.

\bibitem{teterwak2019boundless}
Piotr Teterwak, Aaron Sarna, Dilip Krishnan, Aaron Maschinot, David Belanger,
  Ce Liu, and William~T Freeman.
\newblock Boundless: Generative adversarial networks for image extension.
\newblock In {\em Proceedings of the IEEE International Conference on Computer
  Vision}, pages 10521--10530, 2019.

\bibitem{brown_course_cv}
James Tompkin.
\newblock Brown cs231n: Csci 1430: Introduction to computer vision, spring
  2020.

\bibitem{torralba2011unbiased}
Antonio Torralba and Alexei~A Efros.
\newblock Unbiased look at dataset bias.
\newblock In {\em CVPR 2011}, pages 1521--1528. IEEE, 2011.

\bibitem{trouve2013passive}
Pauline Trouv{\'e}, Fr{\'e}d{\'e}ric Champagnat, Guy Le~Besnerais, Jacques
  Sabater, Thierry Avignon, and J{\'e}r{\^o}me Idier.
\newblock Passive depth estimation using chromatic aberration and a depth from
  defocus approach.
\newblock {\em Applied optics}, 52(29):7152--7164, 2013.

\bibitem{van2019image}
Basile Van~Hoorick.
\newblock Image outpainting and harmonization using generative adversarial
  networks.
\newblock {\em arXiv preprint arXiv:1912.10960}, 2019.

\bibitem{Vorenkamp}
Todd Vorenkamp.
\newblock Understanding crop factor.
\newblock {\em B\&H Explora}, 2016.

\bibitem{wang2019wide}
Yi Wang, Xin Tao, Xiaoyong Shen, and Jiaya Jia.
\newblock Wide-context semantic image extrapolation.
\newblock In {\em Proceedings of the IEEE Conference on Computer Vision and
  Pattern Recognition}, pages 1399--1408, 2019.

\bibitem{willson1994center}
Reg~G Willson and Steven~A Shafer.
\newblock What is the center of the image?
\newblock {\em JOSA A}, 11(11):2946--2955, 1994.

\bibitem{yerushalmy2011digital}
Ido Yerushalmy and Hagit Hel-Or.
\newblock Digital image forgery detection based on lens and sensor aberration.
\newblock {\em International journal of computer vision}, 92(1):71--91, 2011.

\bibitem{zeng2019reliable}
Hui Zeng, Lida Li, Zisheng Cao, and Lei Zhang.
\newblock Reliable and efficient image cropping: A grid anchor based approach.
\newblock In {\em Proceedings of the IEEE Conference on Computer Vision and
  Pattern Recognition}, pages 5949--5957, 2019.

\end{thebibliography}
}

\section*{Supplementary material}

\renewcommand{\thesubsection}{\Alph{subsection}}

\subsection{Dataset constraints, collection, and description} \label{app:dataset}

\subsubsection{Lack of cropping}
As a starting point, any sufficiently large collection of natural photos suffices. In order to simulate scenarios where a user only has access to pixels but not the metadata (which commonly happens when downloading photos from \eg social media), no labels are needed. Training and testing data can be retrieved 'for free' by extracting patches and thumbnails from any dataset consisting of real-world images, where the only important constraint is the lack of tampering. However, it turns out that cropping, as well as various other kinds of 'soft tampering', is a natural part of the digital editing process. Because these operations are mostly harmless and probably happen more often than we realize, it becomes almost impossible to know to what extent a given database really is unedited. \\

\subsubsection{Sufficiently high resolution}
Acknowledging the fact that the dataset might be noisy to some degree, we proceed with adding a resolution constraint. Image datasets for deep learning are often down-scaled such that the maximal dimension lies around 500 to 1,000 pixels\footnote{For example, every sample in Open Images V5 \cite{kuznetsova2018open} has at most 1,024 pixels on its longest side.}, presumably because the benefit of an even finer level of detail for recognizing object semantics rarely outweighs the extra computational cost.
However, in order to better pick up lens flaws that are typically exhibited in subtle pixel-level features, we prefer to keep the resolution higher and closer to the original photo. This matches the observation in image forensics that resizing should be avoided because it tends to damage high-frequency details \cite{marra2020full}. We decided to settle for (\ie download images with) a maximal dimension of 2,048 pixels for each sample, which is deemed high enough to detect optical imperfections, but also low enough to avoid exceeding realistic dimensions of photos that may be shared online. \\

\begin{figure}[t]
    \centering
    \includegraphics[width=\linewidth]{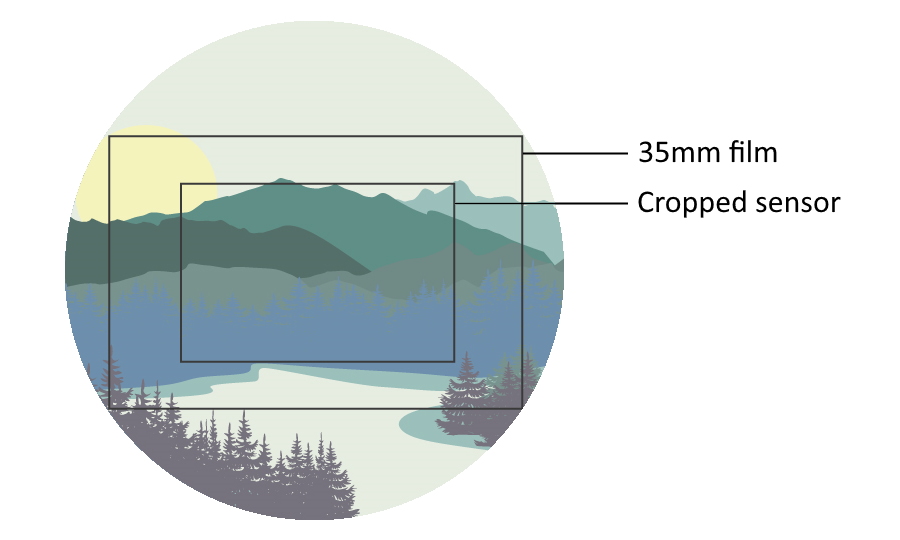}
    \caption{Comparison of a full-frame sensor versus a crop sensor with respect to the lens circle. (Adjusted and reprinted from \cite{Vorenkamp} with permission.)}
    \label{fig:cropfactor}
\end{figure}

\begin{figure*}
    \centering
    \includegraphics[width=\textwidth]{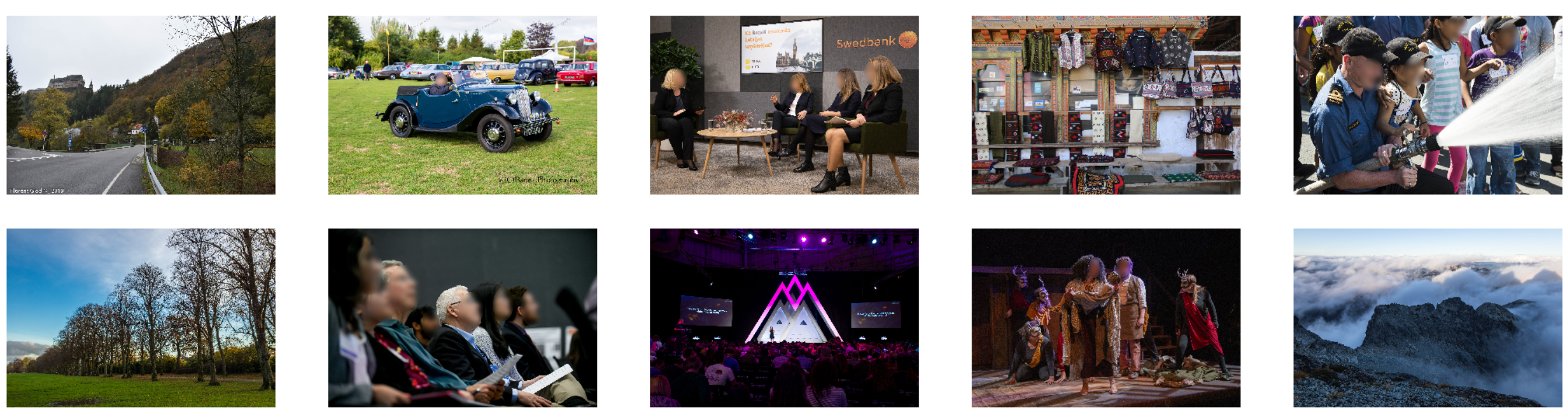}
    \caption{Random examples of the Flickr dataset. (Faced blurred for privacy protection.)}
    \label{fig:flickr_samples}
\end{figure*}

\subsubsection{Inter-device variation considerations}
Every lens and sensor is different, and this variation in standards might make what exactly constitutes a 'crop' less precise. For example, if a full-frame lens is coupled with a crop sensor (\ie the film frame width is less than 35mm) as in Figure \ref{fig:cropfactor}, every resulting picture can be thought of as inherently cropped because the light captured by the sensor does not fully cover the lens circle. Mobile phones have an especially large crop factor, since their sensors are typically much smaller than those used in professional DSLR camera systems.
In fact, there is a vast number of possible configurations, and trying to take all of them into account would become impracticable. We thus clear confusion by defining a 'cropped image' to be any deviation from what was originally captured by the imaging sensor at the time of shooting.
Since our method is camera make and model-blind, we rely on the learning-based approach to discover modal values within this combinatorial space of configurations in the dataset, such that our network will learn to take the diversity among devices and settings into account automatically. \\

\subsubsection{Scraping and dataset bias}
We scraped Flickr by querying the API with 10,000 different search terms and downloading up to 500 photos for every tag. The keywords were gathered from an online list of the 10,000 most commonly used words in English, which was in turn generated by performing N-gram frequency analysis on the Google Trillion Word Corpus \cite{Franz2006,Kaufman2019}.
The resulting database has around 1.3 million images, which would have been 5 million if the search results did not overlap due to many entries having multiple tags. Note that Flickr seems to be biased toward photos (1) depicting persons, (2) of somewhat professional quality, and (3) taken using expensive cameras, but we view neither of these aspects as a drawback considering the relevance of our project to photography patterns and photojournalism. \\

\subsubsection{Aspect ratio}
Mixing training examples with different aspect ratios together also changes the shapes of the grid cells of $F_{patch}$, which should clearly be avoided. Otherwise, patches that have the same absolute position with respect to the lens circle, might be assigned different labels depending on the aspect ratio of the sensor within said lens circle. Most digital camera systems have a sensor size of 36mm$\times$24mm, corresponding to an aspect ratio of 1.5. We therefore fix the aspect ratio to 1.5 and enforce landscape-only photos (by rotating portrait images either left or right) to further enhance consistency, which shrinks the pool of files meeting all discussed criteria down to 700,000 files.

\subsubsection{Dataset split}
Lastly, we perform a 3-way train / validation / test set split distributed as 90\% / 5\% / 5\%. A few samples of the test set are shown in Figure \ref{fig:flickr_samples}.

\subsection{Shortcut mitigation} \label{app:shortcut_fuzzing}

Convolutional neural networks have been shown to be surprisingly adept at finding and leveraging often irrelevant shortcuts \cite{doersch2015unsupervised, noroozi2017representation}. Here, we present our approach to ensure that the models learn useful features.

\subsubsection{Image patch extraction}

Patches are extracted from the centers of a regularly sized $4\times4$ grid within every image (cropped or not), but we also apply random jittering of $\pm8$ pixels in both dimensions.
This way, we discourage $F_{patch}$ from learning low-level image processing-related shortcuts, for example JPEG block artefact alignment.

\subsubsection{Resizing global images}

Since $F_{global}$ uses a downscaled variant of the incoming image with fixed dimensionality $224\times149$, but cropping an image also changes its raw dimensions, we were obliged to employ some tricks in order to prevent the model from learning glitches that are unrelated to physical imaging aberrations, notably resampling factor detection.
Resampling shortcuts have occurred in various previous works \cite{noroozi2017representation}, and are typically an undesired factor.
For example, a neural network is able to trivially distinguish images that have been downsized starting from $2048\times1365$ as opposed to starting from $1536\times1024$ based on pixel-level resampling artefacts, even if the interpolation method is randomized \cite{noroozi2017representation}.
To work around this issue, we perform random resizing in multiple stages to make the original dimensions nearly impossible to recover, without noticeably damaging the image contents.

Given the potentially cropped source image of size $W \times H$, we first resize 3 times to a random $W' \times H'$ where $W'$ is uniformly distributed in $[1024, 2048]$, and $H'|W'$ is conditionally uniformly distributed in $[0.8W'/A, 1.2W'/A]$, with $A$ the aspect ratio. Note that the interpolation method itself is also random, and is chosen from one of \{NEAREST, LINEAR, AREA, CUBIC, LANCZOS4\} as provided by the OpenCV library \cite{opencv}. Finally, the whole image is downscaled to $224\times149$, and from now on it should be nearly impossible to tell what its original resolution was.

Indeed, if we replace the cropping operation with a rescaling to the same dimensions that the cropped image would otherwise have, the accuracy of our global model drops to chance ($50\%$). This suggests that only altered image contents play a role, while input resolution does not anymore.

Note that the way in which patches are extracted remains unaltered by this procedure; only thumbnails must be treated to ensure that $F_{global}$ predominantly looks at semantically meaningful content.

\subsubsection{Joint model}

Another, more sophisticated shortcut arose which occurs only when the model has access to both patches and thumbnails simultaneously.
Even if the original dimensions of a global image cannot be inferred, the integrated network could still learn to measure how 'large' the patches are in comparison to the thumbnail, since they are extracted from a 'smaller' image if the input is cropped.
To alleviate this issue, we perform an extra random resizing step \textit{before} extracting patches but \textit{after} cropping, where the width is uniformly distributed in $[1024, 2048]$ and the height is chosen proportionally such that the aspect ratio is retained. This guarantees that the fraction of the thumbnail that is being covered by patches loses its predictive power, discouraging $G$ from trying to exploit low-level correlations among the outputs produced by $F_{patch}$ and $F_{global}$.
This approach serves the additional purpose of enforcing our ignorance about both the crop rectangle and the sequence of resizes that images at test time could have undergone; hence, during our evaluations, we also randomize input resolutions the same way.

\subsubsection{Patch labels and intra-batch interaction}

We observed a peculiar effect when all the examples within a minibatch have the same ground truth label for absolute patch localization. Specifically, when all patches belonging to the same position class were forwarded through the network, an unnaturally high accuracy could be achieved during training, but not during validation. This does not occur when the batch size is just 1 instead of 64, implying that there exists an architectural feature of the neural network that enables cross-example interaction. Although we have not studied this aspect systematically, we speculate that it may be due to the BatchNorm2D layers contained in a ResNet \cite{he2016deep}, whereby the mutual information across different examples within every minibatch is somehow leaked and exploited. To counter this shortcut, we cyclically shift image patches across minibatches in order to ensure that every minibatch contains a uniform distribution of all 16 labels, rather than all of them having the same class. The mutual information among examples within a minibatch is therefore minimized, and the peculiar overfitting effect disappeared, as evidenced by the results becoming independent of batch size.

\begin{figure}
    \centering
    \includegraphics[width=\linewidth]{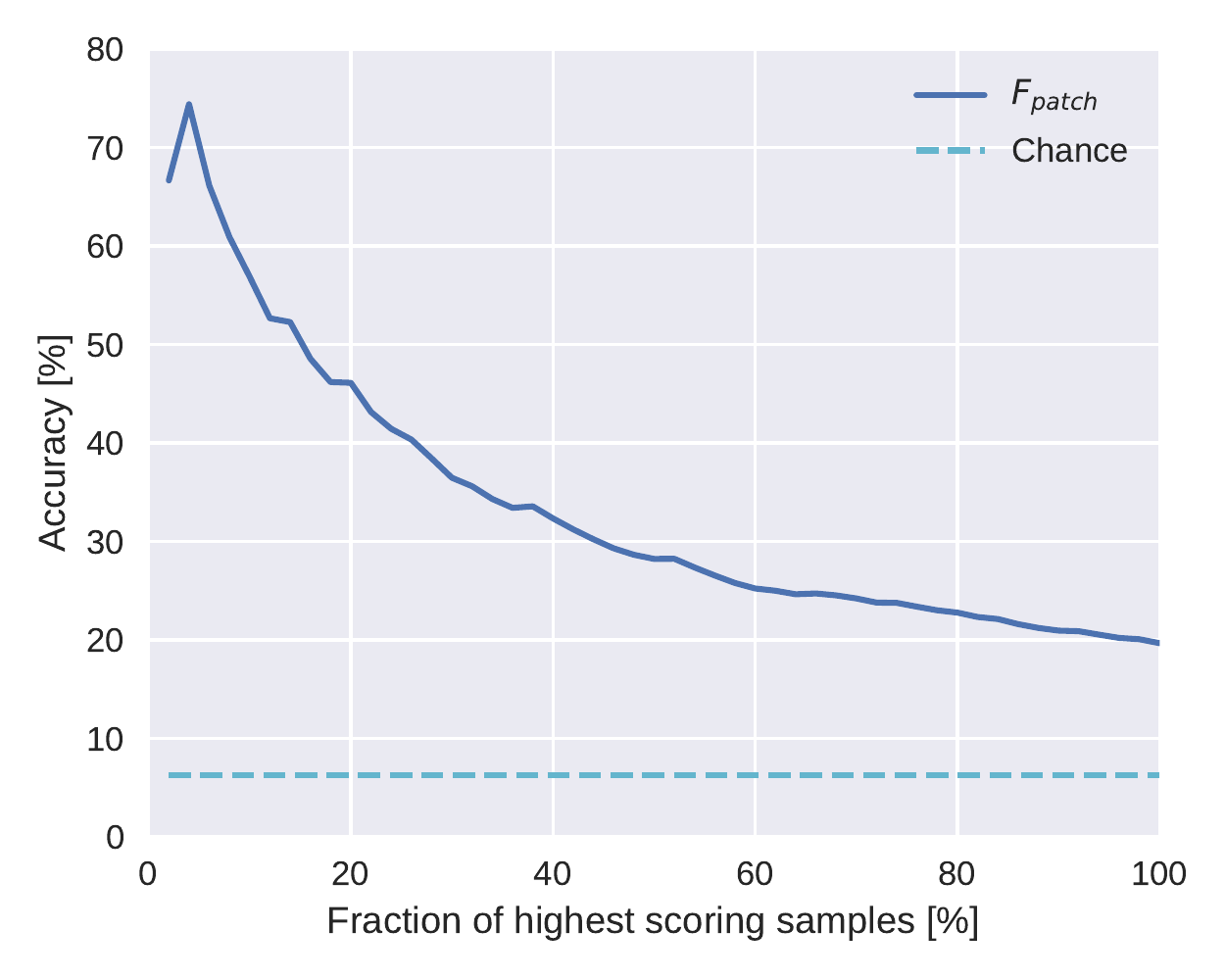}
    \caption{\textbf{Sample selectivity versus patch localization performance.} The accuracy improves significantly once we discard more and more predictions that $F_{patch}$ is uncertain about.}
    \label{fig:patch_sel}
\end{figure}

\begin{figure}
    \centering
    \begin{subfigure}{0.48\linewidth}
        \centering
        \includegraphics[width=\linewidth]{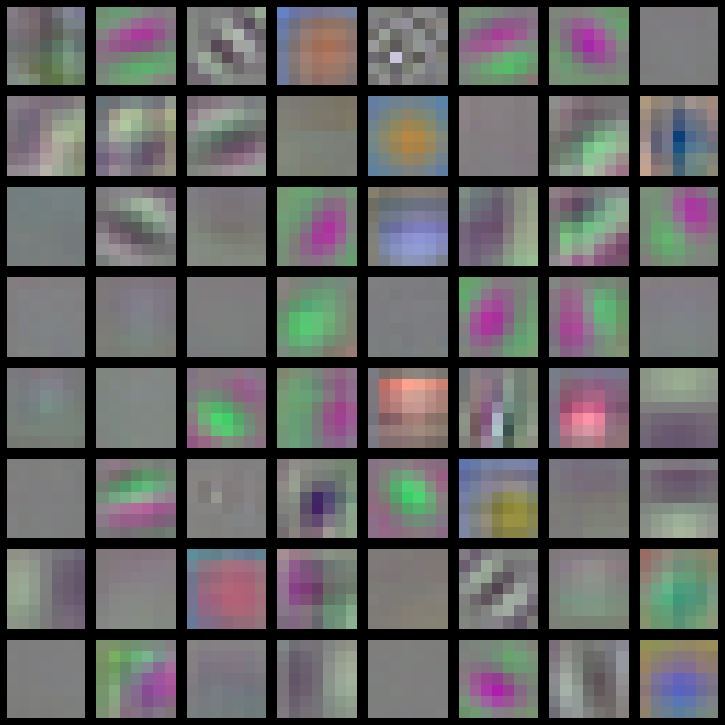}
        \caption{$F_{patch}$ (ResNet-18).}
    \end{subfigure}%
    ~
    \begin{subfigure}{0.48\linewidth}
        \centering
        \includegraphics[width=\linewidth]{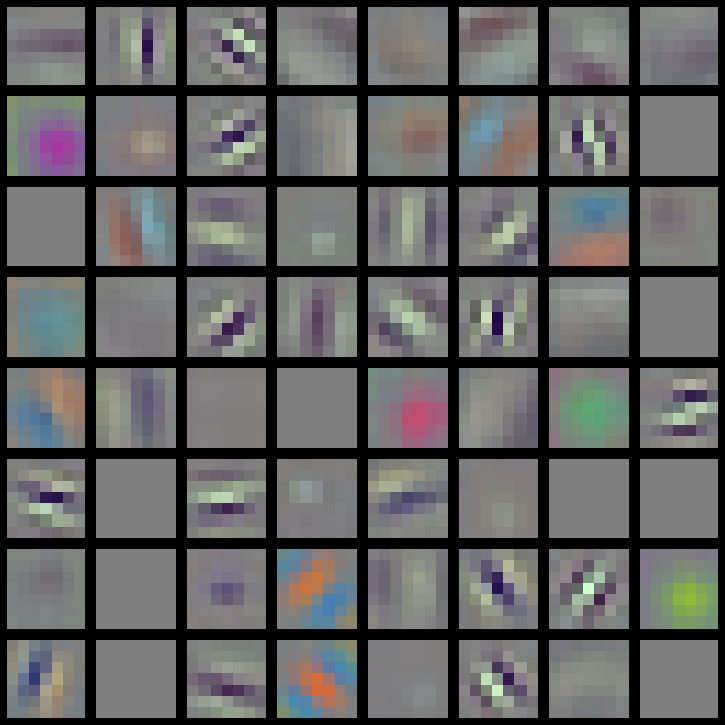}
        \caption{ImageNet-trained ResNet-34.}
    \end{subfigure}
    \caption{\textbf{First convolutional layer filter visualization.} At the lowest level, the absolute patch localization model is clearly more sensitive to alternations between green and magenta (\ie lack of green) pixel values in various directions, as compared to a vanilla ImageNet-trained neural network.}
    \label{fig:conv1}
\end{figure}

\subsection{Patch localization accuracy versus confidence}

Figure \ref{fig:patch_sel} plots the accuracy of $F_{patch}$ as a function of the response rate, where moving to the left on the horizontal axis means that an increasingly smaller fraction of only the patches with the highest scores are considered. This supports the earlier claim that the maximum value in the output distribution correlates positively with the correctness of the pretext model.

\subsection{Convolutional filter visualization}

We display and compare the values of the convolution operations applied by the very first layers of both $F_{patch}$ and a regular ImageNet classifier in Figure \ref{fig:conv1}. These visualizations suggest that the network is particularly sensitive to green transverse chromatic aberration. 

\begin{figure*}
    \centering
    \begin{subfigure}{\textwidth}
        \centering
        \includegraphics[width=\textwidth]{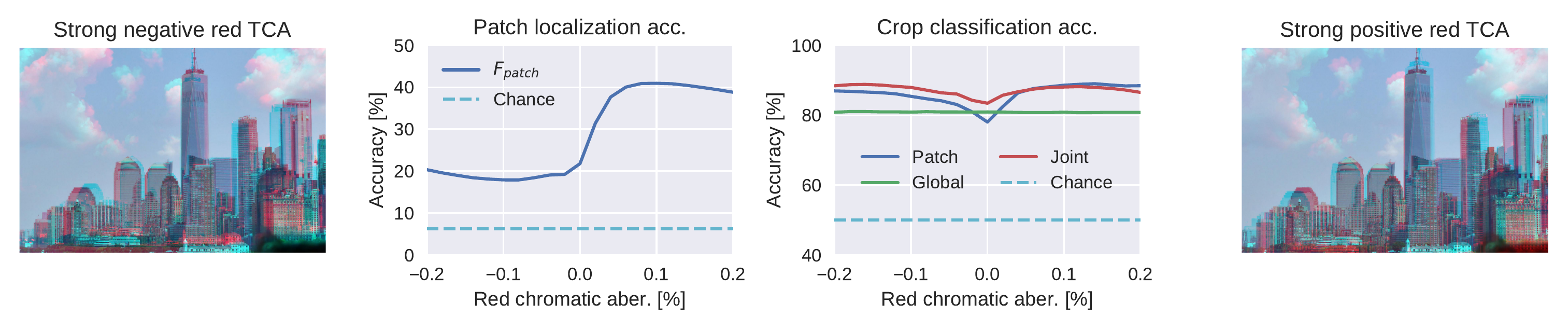}
        \caption{\textbf{Red transverse chromatic aberration} in the positive (outward) direction boosts performance.}
        \label{fig:sweep_rca}
    \end{subfigure}
    \par\medskip
    \begin{subfigure}{\textwidth}
        \centering
        \includegraphics[width=\textwidth]{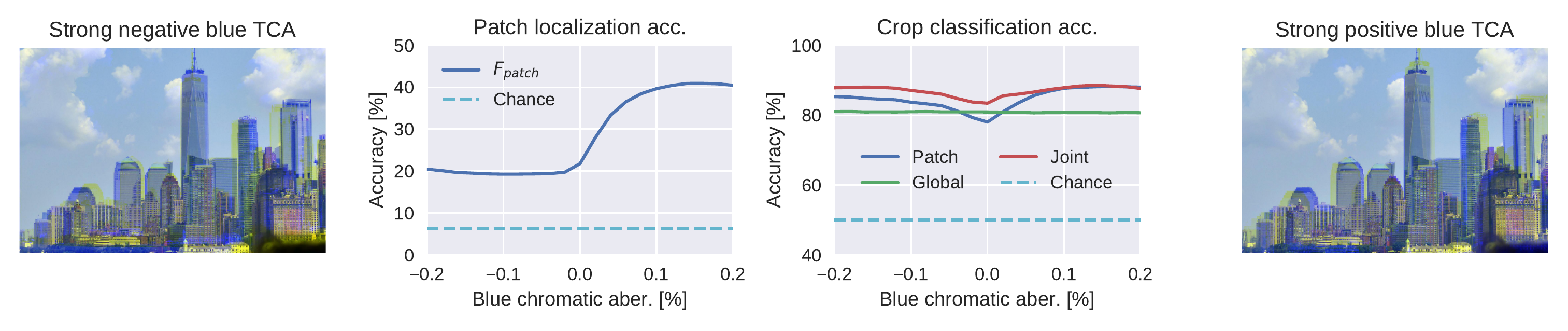}
        \caption{\textbf{Blue transverse chromatic aberration} in the positive (outward) direction boosts performance.}
        \label{fig:sweep_bca}
    \end{subfigure}
    \par\medskip
    \begin{subfigure}{\textwidth}
      \centering
      \includegraphics[width=\textwidth]{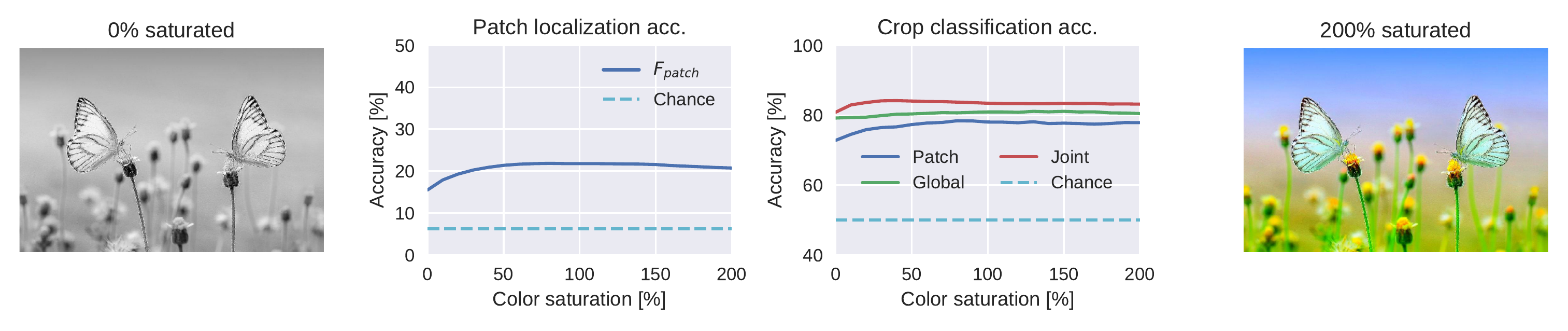}
      \caption{Adjusting \textbf{color saturation} away from 100\% (= identity) slightly degrades performance.}
      \label{fig:sweep_sat}
    \end{subfigure}
    \par\medskip
    \begin{subfigure}{\textwidth}
      \centering
      \includegraphics[width=\textwidth]{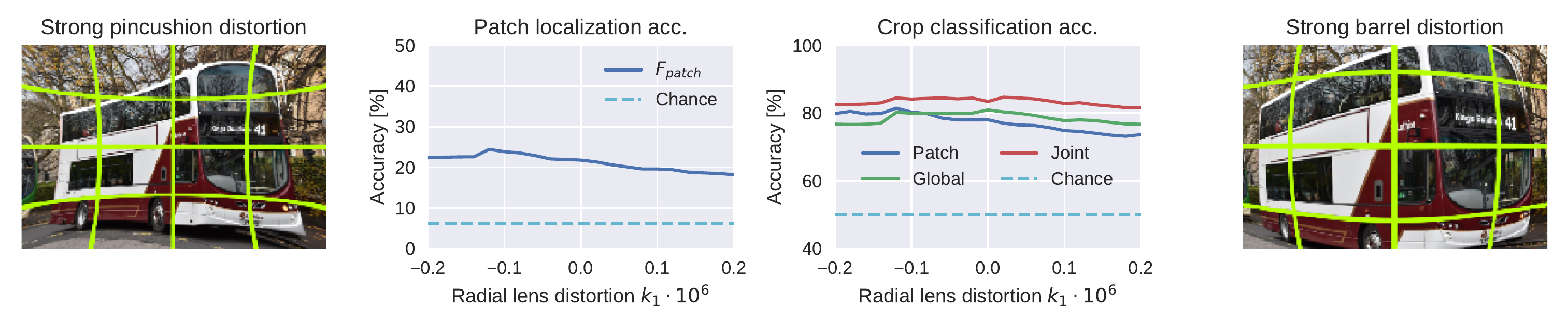}
      \caption{The degree of \textbf{radial lens distortion} in our dataset may be too subtle to substantially affect the integrated crop detection model, although due to the noisy results, this is inconclusive.}
      \label{fig:sweep_rd}
    \end{subfigure}
    \caption{\textbf{Extended breakdown of image attributes.} See Figure \ref{fig:sweep} for the main results.}
    \label{fig:more_sweep}
\end{figure*}

\begin{table}
    \centering
    \begin{tabular}{l r r r}
        \toprule
        \textbf{Model} & \textbf{Color} & \textbf{Grayscale} & \textbf{Chance} \\ \midrule
        $F_{patch}$ (patch loc.) & 21\% & 15\% & 6\% \\ \midrule
        Joint (crop det.) & 86\% & 81\% & 50\% \\
        Global (crop det.) & 79\% & 78\% & 50\% \\
        Patch (crop det.) & 77\% & 72\% & 50\% \\
        \bottomrule
    \end{tabular}
    \caption{\textbf{Accuracies with or without color.} Removing all color information on the test set decreases the model's performance, but only considerably so when a model relies on patches.}
    \label{tab:sat_gray}
\end{table}

\subsection{Additional experiments for lens-related clues} \label{app:more_clues}

\subsubsection{Effect of red and blue chromatic aberration}

As shown in Figures \ref{fig:sweep_rca} and \ref{fig:sweep_bca}, the patch localization accuracy plots appear horizontally flipped with respect to Figure \ref{fig:sweep_gca}.
This indicates that the modal value of purple fringing in our dataset corresponds to the green channel being scaled toward one preferred direction more often than in the other direction. (Inward green TCA is visually the same as a combination of outward red and blue TCA.)

\subsubsection{Effect of color saturation and grayscale}

In order to quantify the significance of color information in general beyond just chromatic aberration, it may be instructive to control the saturation of the test set. A saturation factor of 0\% is equivalent to grayscale imagery, 100\% is identity, and larger numbers represent exaggerated colors. The result is shown in Figure \ref{fig:sweep_sat}. This feature does not depend on the location of a patch, therefore it is not unexpected that the best performance corresponds with untouched images. Any other value simply moves the images away from the expected distribution.

Table \ref{tab:sat_gray} also compares the performance of the model when tested on grayscale and regular color images.
Although color information clearly constitutes a respectable gain to the network's correctness relative to chance levels, there is a large residual gap that does not rely on color. Apart from vignetting, we hypothesize this is mostly related to photography patterns and object priors, which we discussed in Section \ref{sec:photobias}.
Moreover, the only model that is likely unable to perceive lens aberrations in the first place ($\textit{global}$) seems to care the least about color information, suggesting that the object priors involved in revealing crops can be learned with minimal dependence on color.

\subsubsection{Effect of radial lens distortion}

Pincushion or barrel distortion, illustrated left and right respectively in Figure \ref{fig:sweep_rd}, arises from the fact that the magnification of a scene through a lens does not stay constant across the image plane, but depends on the radius $r=\sqrt{x^2+y^2}$ from the optical center \cite{lin2018image}. We replicate this distortion by applying a geometric coordinate transformation with a simple square law that scales every destination pixel $(x_d,y_d)$ relative to its source $(x_s,y_s)$ as follows:
\begin{align}
d &= 1+k_1r^2 \label{eq:dist1} \\
(x_d,y_d) &= (dx_s,dy_s) \label{eq:dist2}
\end{align}
Figure \ref{fig:sweep_rd} shows the effect of inflating lens distortion on the test set according to Equation (\ref{eq:dist1}-\ref{eq:dist2}).

\end{document}